\documentclass[10pt,journal,compsoc]{IEEEtran}

\usepackage{graphicx}
\usepackage[section]{algorithm}
\usepackage[noend]{algorithmic}
\usepackage{amsmath}
\usepackage{bm}
\usepackage{color}
\usepackage{ragged2e}
\usepackage{subfigure}
\usepackage{url}
\usepackage[numbers,sort&compress]{natbib}

\renewcommand{\raggedright}{\leftskip=0pt \rightskip=0pt plus 0cm}

\usepackage{array}
\newcommand{\PreserveBackslash}[1]{\let\temp=\\#1\let\\=\temp}
\newcolumntype{C}[1]{>{\PreserveBackslash\centering}p{#1}}
\newcolumntype{R}[1]{>{\PreserveBackslash\raggedleft}p{#1}}
\newcolumntype{L}[1]{>{\PreserveBackslash\raggedright}p{#1}}

\begin{document}

\title{A Domain Adaptive Density Clustering Algorithm for Data with Varying Density Distribution}

\author{Jianguo~Chen, ~\IEEEmembership{Member, IEEE}
        and Philip S. Yu, ~\IEEEmembership{Fellow, IEEE}
\IEEEcompsocitemizethanks{
\IEEEcompsocthanksitem Jianguo~Chen is with the College of Computer Science and Electronic Engineering, Hunan University, Changsha, Hunan 410082, China (jianguochen@hnu.edu.cn).
\IEEEcompsocthanksitem Philip S. Yu is with the Department of Computer Science, University of Illinois at Chicago, Chicago, IL 60607, USA, and Institute for Data Science, Tsinghua University, Beijing 100084, China (psyu@uic.edu).}
}

\markboth{}
{J. Chen \MakeLowercase{\textit{et al.}}: IEEE Transactions on Knowledge and Data Engineering}

\IEEEtitleabstractindextext{
\begin{abstract}
 \raggedright{
As one type of efficient unsupervised learning methods, clustering algorithms have been widely used in data mining and knowledge discovery with noticeable advantages.
However, clustering algorithms based on density peak have limited clustering effect on data with varying density distribution (VDD), equilibrium distribution (ED), and multiple domain-density maximums (MDDM), leading to the problems of sparse cluster loss and cluster fragmentation.
To address these problems, we propose a Domain-Adaptive Density Clustering (DADC) algorithm, which consists of three steps:  domain-adaptive density measurement, cluster center self-identification, and cluster self-ensemble.
For data with VDD features, clusters in sparse regions are often neglected by using uniform density peak thresholds, which results in the loss of sparse clusters.
We define a domain-adaptive density measurement method based on $K$-Nearest Neighbors (KNN) to adaptively detect the density peaks of different density regions.
We treat each data point and its KNN neighborhood as a subgroup to better reflect its density distribution in a domain view.
In addition, for data with ED or MDDM features, a large number of density peaks with similar values can be identified, which results in cluster fragmentation.
We propose a cluster center self-identification and cluster self-ensemble method to automatically extract the initial cluster centers and merge the fragmented clusters.
Experimental results demonstrate that compared with other comparative algorithms, the proposed DADC algorithm can obtain more reasonable clustering results on data with VDD, ED and MDDM features.
Benefitting from a few parameter requirement and non-iterative nature, DADC achieves low computational complexity and is suitable for large-scale data clustering.
}
\end{abstract}

}

\maketitle
\IEEEdisplaynontitleabstractindextext
\IEEEpeerreviewmaketitle

\section{Introduction}
\IEEEPARstart{C}{lustering} algorithms have been widely used in various data analysis fields \cite{ex91, ex90}.
Numerous clustering algorithms have been proposed, including the partitioning-based, hierarchical-based, density-based, grid-based, model-based, and density-peak-based methods \cite{ex08,ex88, ex09, ex10}.
Among them, density-based methods (e.g., DBSCAN, CLIQUE, and OPTICS) can effectively discover clusters of arbitrary shape using the density connectivity of clusters, and do not require a pre-defined number of clusters \cite{ex10}.
In recent years, Density-Peak-based Clustering (DPC) algorithms, as a branch of density-based clustering, were introduced in \cite{ex13, ex31}, assuming that the cluster centers are surrounded by low-density neighbors and can be detected by efficiently searching for local density peaks.

\begin{figure}[!ht]
  \centering
  \includegraphics[width=3.2in]{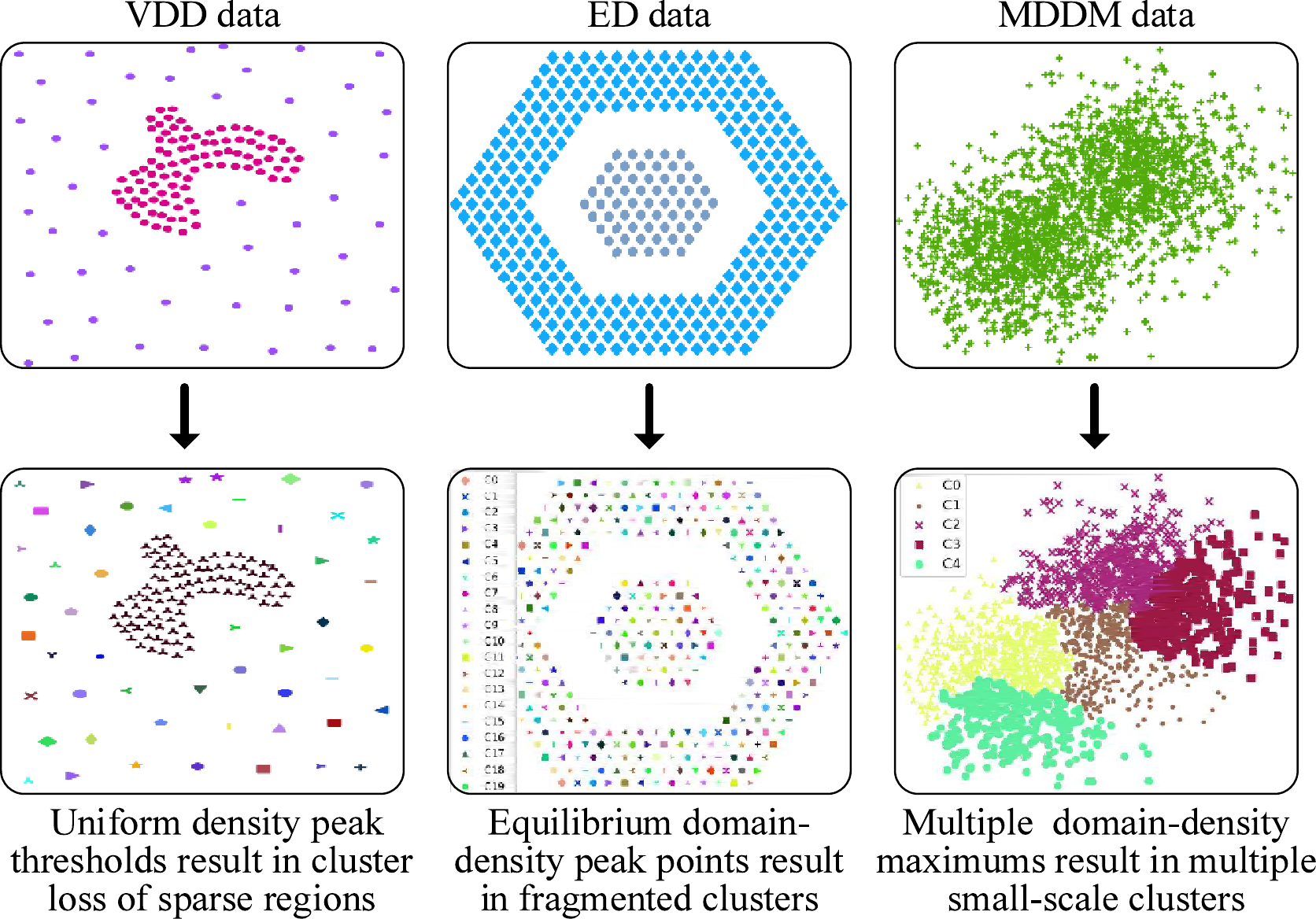}
  \caption{Challenges of DPC algorithms on data with VDD, ED, and MDDM.}
  \label{img001}
\end{figure}

Benefitting from few parameter requirements and non-iterative nature, DPC algorithms can efficiently detect clusters of arbitrarily shape from large-scale datasets with low computational complexity.
However, as shown in Fig. \ref{img001}, DPC algorithms have limited clustering effect on data with varying density distribution (VDD), multiple domain-density maximums (MDDM), or equilibrium distribution (ED).
(1) For data with VDD characteristics, there are varying density regions and data points in sparse regions are usually ignored as outliers or misallocated to adjacent dense clusters by using uniform density peak thresholds, which results in the loss of sparse clusters.
(2) Clustering results of DPC algorithms depend on a strict constraint that there is only one local density maximum in each candidate cluster.
However, for data with MDDM or ED, there are zero or more local density maximums in a natural cluster, and DPC algorithms might lead to the problem of cluster fragmentation.
(3) In addition, how to determine the parameter thresholds of local density and Delta distance in the clustering decision graph is another problem for DPC algorithms.
Therefore, it is critical to address the problems of sparse cluster loss and cluster fragmentation from data with VDD, ED, and MDDM and improve clustering accuracy.

\begin{figure}[!ht]
  \centering
  \includegraphics[width=3.4in]{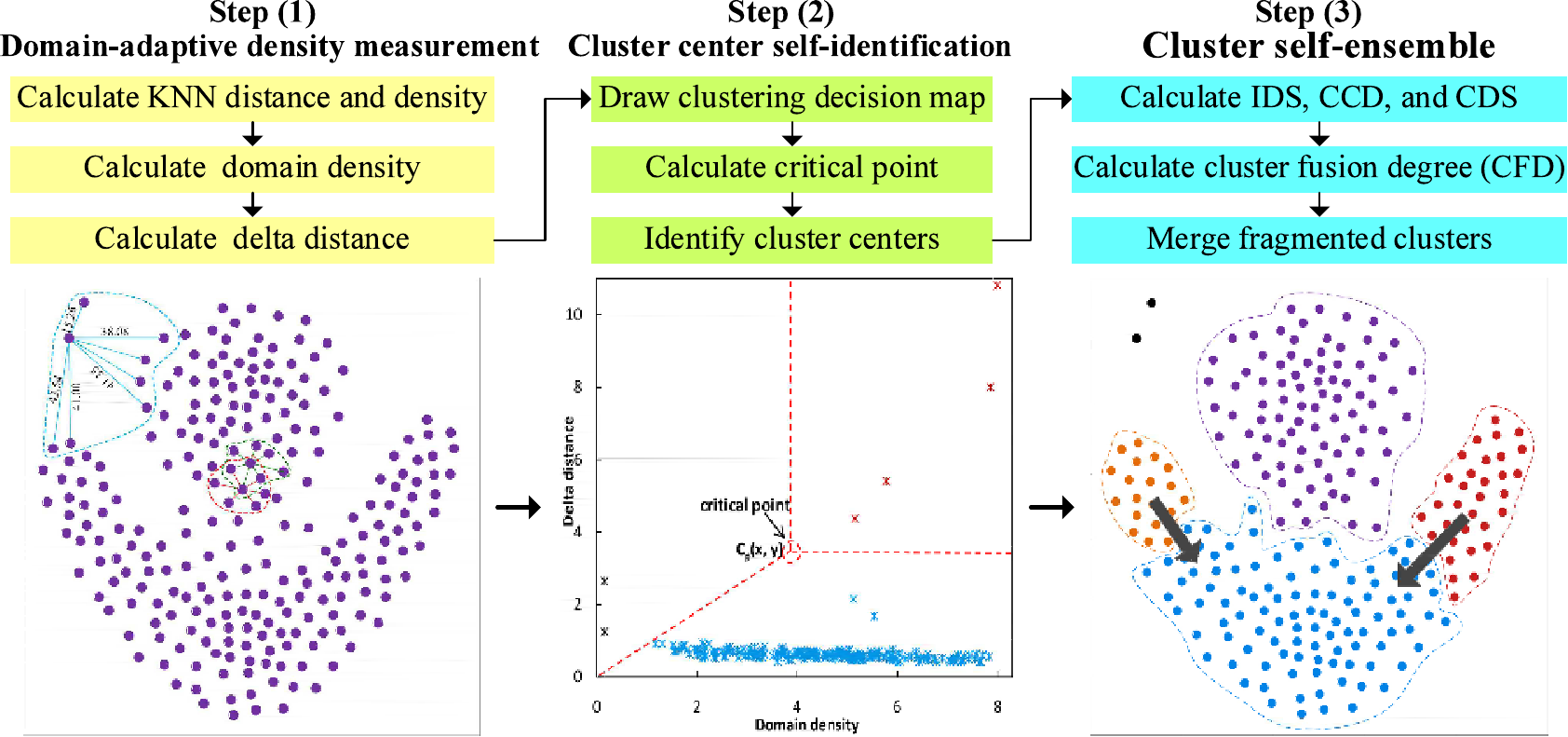}
  \caption{Workflow of the proposed DADC algorithm.}
  \label{img002}
\end{figure}

Aiming at the problems of sparse cluster loss and cluster fragmentation, we propose a Domain-Adaptive Density Clustering (DADC) algorithm.
As shown in Fig. \ref{img002}, the DADC algorithm consists of three steps: domain-adaptive density measurement, cluster center self-identification, and cluster self-ensemble.
A domain-adaptive density measurement method based on K-Nearest Neighbors (KNN) is defined, which can be used to adaptively detect the density peaks of different density regions.
On this basis, cluster center self-identification and cluster self-ensemble methods are proposed to automatically extract the initial cluster centers and merge the fragmented clusters.
Extensive experiments indicate that DADC outperforms comparison algorithms in clustering accuracy and robustness.
The contributions of this paper are summarized as follows.

\begin{itemize}
  \item To address the problem of sparse cluster loss of data with VDD, a domain-adaptive density measurement method is proposed to detect density peaks in different density regions.
      According to these density peaks, cluster centers in both dense and sparse regions are effectively discovered, which well addresses the sparse cluster loss problem.
  \item To automatically extract the initial cluster centers, we draw a clustering decision graph based on domain density and Delta distance.
      We then propose a cluster center self-identification method and automatically determine the parameter thresholds and cluster centers from the clustering decision graph.
  \item To address the problem of cluster fragmentation on data with ED or MDDM, an innovative Cluster Fusion Degree (CFD) model is proposed, which consists of the inter-cluster density similarity, cluster crossover degree, and cluster density stability.
      Then, a cluster self-ensemble method is proposed to automatically merge the fragmented clusters by evaluating the CFD between adjacent clusters.
\end{itemize}

The rest of the paper is organized as follows.
Section \ref{section2} reviews the related work.
Section \ref{section3} presents the domain adaptive method for cluster center detection.
A cluster self-identification method and cluster ensemble method are respectively introduced in Section \ref{section4}.
Experimental results and evaluations are shown in Section \ref{section5}.
Finally, Section \ref{section6} concludes the paper.

\section{Related Work}
\label{section2}
Being an efficient and unsupervised data mining method, numerous clustering algorithms are proposed and widely applied in various applications \cite{ex05,ex90, ex96}.
Partition-based methods (e.g., K-Means and K-Medoids) \cite{ex08} are easy to understand and implement, but it is sensitive to noisy data and can only detect round or spherical clusters.
Hierarchical methods (e.g., BIRCH, CURE, and ROCK) \cite{ex09} do not need to pre-define a number of clusters and can extract hierarchical relationship of clusters, but require high computational complexity.
Density-based methods (e.g., DBSCAN, CLIQUE, and OPTICS) \cite{ex10} also don't require a pre-defined number of clusters and can discover clusters of arbitrary shapes, but the clustering results of these algorithms are sensitive to the threshold of their parameters.

Focusing on density-based clustering analysis, abundant improvements of traditional algorithms were presented, while novelty algorithms were explored \cite{ex13, ex18, ex16, ex93}.
Groups of Density-Peak-based Clustering (DPC) algorithms were proposed in \cite{ex13, ex31}, where cluster centers are detected by efficiently searching of density peaks.
In \cite{ex13}, Rodriguez \emph{et al}. proposed a DPC algorithm titled ``Clustering by fast search and find of density peaks'' (widely quoted as CFSFDP).
CFSFDP can effectively detect arbitrarily shaped clusters from large-scale datasets.
Benefiting from non-iterative nature, CFSFDP achieves low computational complexity and high efficiency for big data processing.
In addition, considering large-scale noisy datasets, robust clustering algorithms were discussed in \cite{ex19} by detecting density peaks and assigning points based on fuzzy weighted KNN method.

For data that exhibit varying-density distribution or multiple local-density maximums, DPC algorithms face a variety of limitations, such as sparse cluster loss and cluster fragmentation.
To address these problems, a variety of optimization solutions were presented in \cite{ex25, ex26, ex27}.
Zheng \emph{et al}. proposed an approximate nearest neighbor search method for multiple distance functions with a single index \cite{ex25}.
To overcome the limitations of DPC, an adaptive method was presented in \cite{ex40} for clustering, where heat-diffusion is used to estimate density and cutoff distance is simplified.
In \cite{ex39}, an adaptive density-based clustering algorithm was introduced in spatial databases with noise, which uses a novel adaptive strategy for neighbor selection based on spatial object distribution to improve clustering accuracy.

Aiming at clustering ensemble, an automatic clustering approach was introduced via outward statistical testing on density metrics in \cite{ex26}.
A nonparametric Bayesian clustering ensemble method was explored in \cite{ex82} to seek the number of clusters in consensus clustering, which achieves versatility and superior stability.
Yu \emph{et al}. proposed an adaptive ensemble framework for semi-supervised clustering solutions \cite{ex27}.
Zeng \emph{et al}. proposed a framework for hierarchical ensemble clustering \cite{ex09}.
Yu \emph{et al}. introduced an incremental semi-supervised clustering ensemble approach for high-dimensional data clustering \cite{ex29}.

Compared with the existing clustering algorithms, the proposed domain-adaptive density method in this work can adaptively detect the domain densities and cluster centers in regions with different densities.
This method is very feasible and practical in actual big data applications.
The proposed cluster self-identification method can effectively identify the candidate cluster centers with minimum artificial intervention.
Moreover, the proposed CFD model takes full account of the relationships between clusters of large-scale datasets, including the inter-cluster density similarity, cluster crossover degree, and cluster density stability.

\section{Domain-Adaptive Density Method}
\label{section3}
We propose a domain-adaptive density method to address the problem of spare cluster loss of DPC algorithms on VDD data.
Domain density peaks of data points in regions with different densities are adaptively detected.
In addition, candidate cluster centers are identified based on the decision parameters of the domain densities and Delta distances.

\subsection{Problem Definitions}
\label{section3.1}
Most DPC algorithms \cite{ex13,ex31} are based on the assumption that a cluster center is surrounded by neighbors with lower local densities and has great Delta distances from any relative points with higher densities.
For each data point $x_{i}$, its local density $\rho_{i}$ and Delta distance $\delta_{i}$ are calculated from the higher density points.
These two quantities depend only on the distances between the data points.
The local density $\rho_{i}$ of $x_{i}$ is defined as:

\begin{equation}
\label{eq01}
\rho_{i} = \sum_{j}{\chi (d_{ij} - d_{c})},
\end{equation}
where $d_{c}$ is a cutoff distance and $\chi (x) = 1$, if $x < 0$; otherwise, $\chi (x) = 0$.
Basically, $\rho_{i}$ is equal to the number of points closer than $d_{c}$ to $x_{i}$.
Delta distance $\delta_{i}$ of $x_{i}$ is measured by computing the shortest distance between $x_{i}$ and any other points with a higher density, as defined as:
\begin{equation}
\label{eq02}
\delta_{i} = \min_{j:\rho_{j} > \rho_{i}}{d_{ij}}.
\end{equation}
For the highest density point, $\delta_{i} = \max_{j}{d_{ij}}$.
Points with a high $\rho$ and high $\delta$ are considered as cluster centers, while points with a low $\rho$ and a high $\delta$ are considered as outliers.
After finding the cluster centers, each remaining point is assigned to the same cluster as its nearest neighbor of higher density.

Most data in actual applications have the characteristics of noise, irregular distribution, and sparsity.
In particular, the density distribution of data points is unpredictable and discrete in most of the cases.
For a VDD dataset, there coexist regions with different degrees of density, such as dense and sparse regions, as defined as follows.

\textbf{Definition 3.1}
\textit{VDD Data.
For a dataset has multiple regions, the average density of data points in each region is set as the region's density.
If there coexist regions with obvious different region densities, such as dense and sparse regions, we denote the dataset as a Varying-Density Distributed (VDD) dataset.}

\begin{figure}[!ht]
  \centering
  \subfigure[Data points]{\includegraphics[width=1.6in]{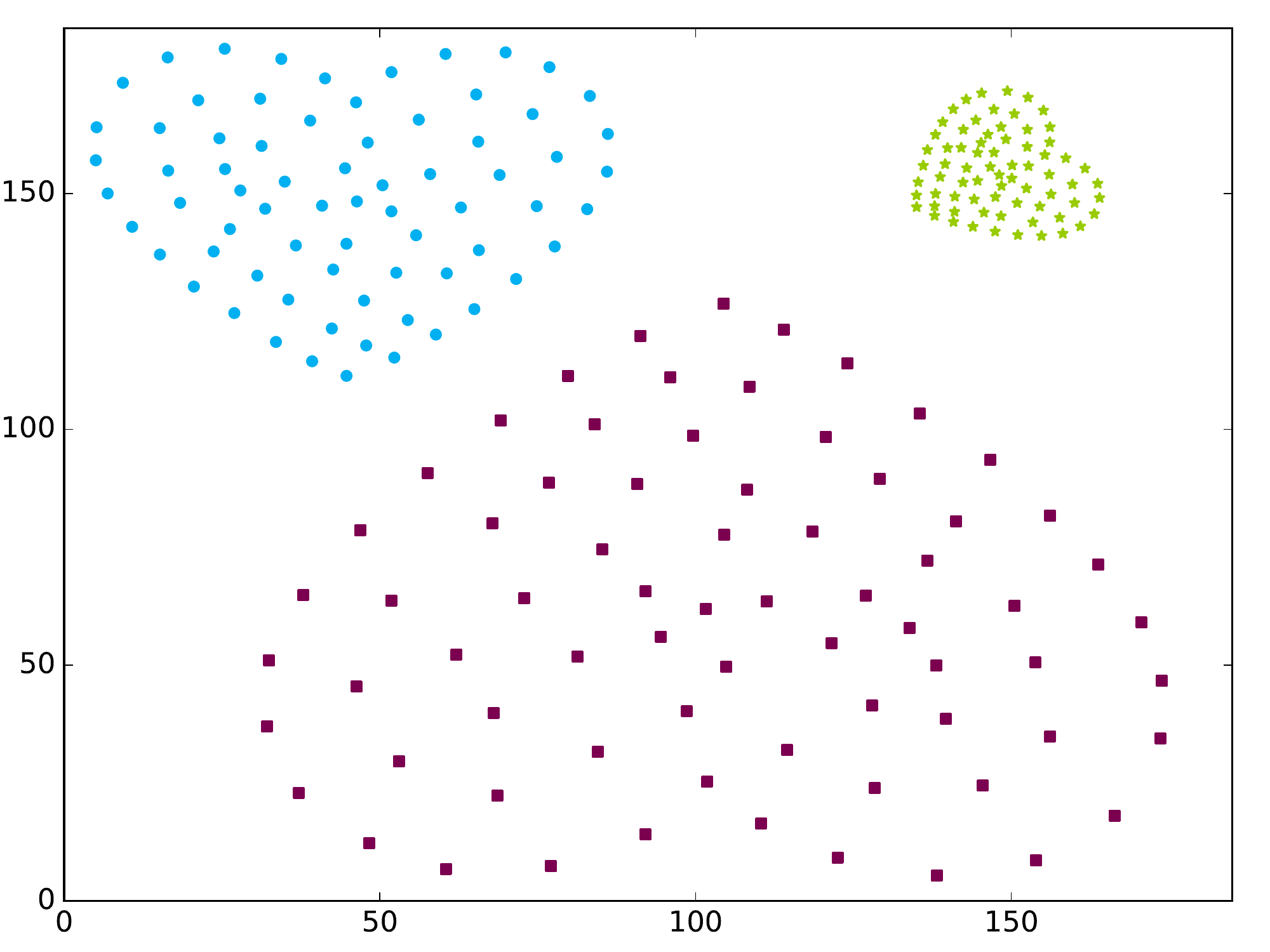}}
  \subfigure[Decision graph]{ \includegraphics[width=1.5in]{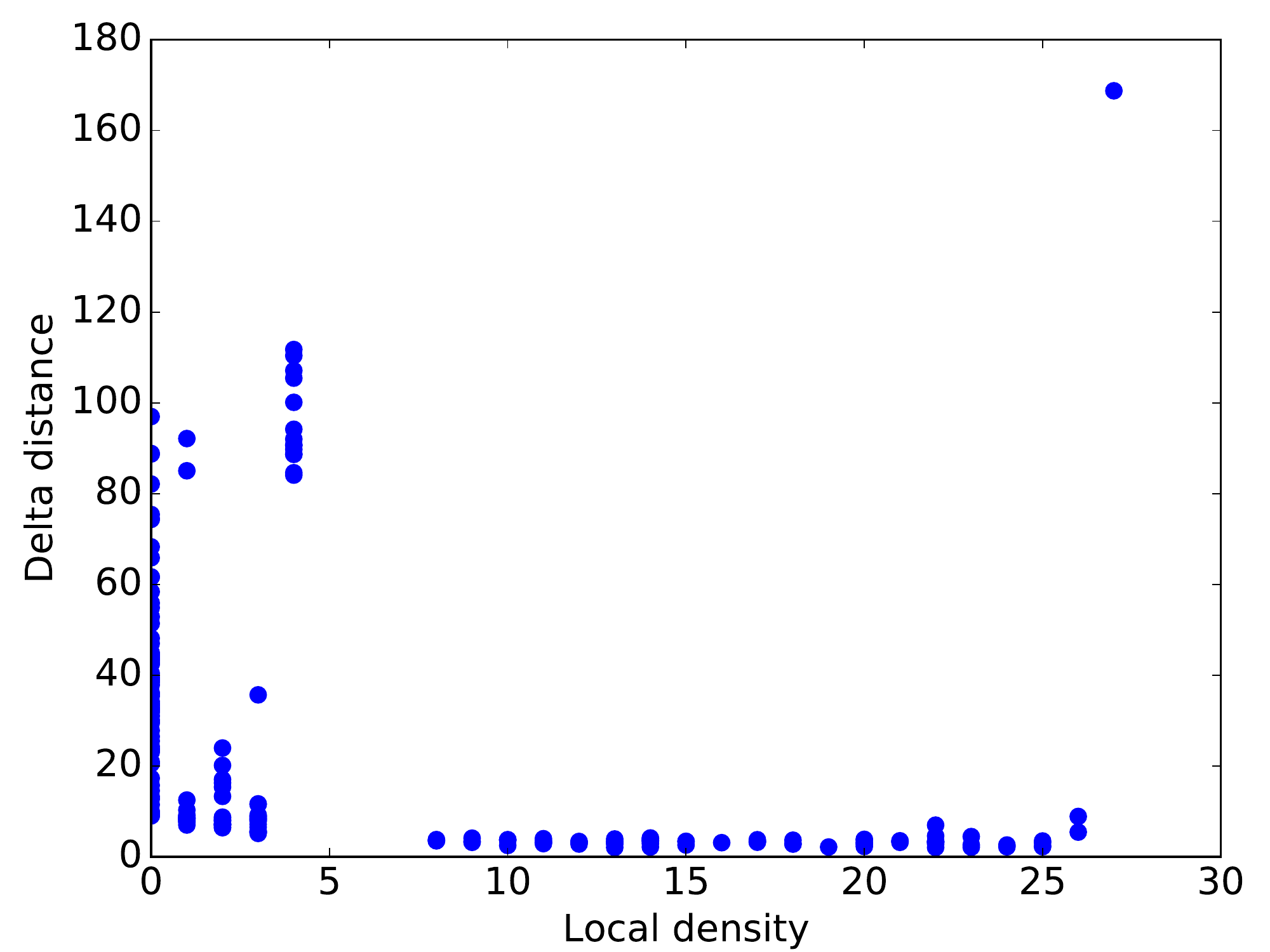}}
  \caption{{\small Example of the CFSFDP algorithm on a VDD dataset.}}
  \label{img01}
\end{figure}

The CFSFDP algorithm and other DPC algorithms suffer from the limitations of sparse cluster loss on VDD datasets.
According to Eq. (\ref{eq01}), points in the relatively sparse area are easily ignored as outliers.
An example of the CFSFDP clustering results on a VDD dataset is shown in Fig. \ref{img01}.

In Fig. \ref{img01} (a),  the heart-shaped dataset has three regions with different densities.
The clustering decision graph achieved by CFSFDP is shown in Fig. \ref{img01} (b), where only one point is obtained with high values of both $\rho$ and $\delta$.
Consequently, the dataset is clustered into one cluster, while the data points in the sparse regions, indicated by blue dots and purple squares, are removed as outliers or incorporated into the dense cluster.

\subsection{Domain-Adaptive Density Measurements}
\label{section3.2}
To adaptively detect the domain-density peaks in different density areas of VDD data, a domain-adaptive density calculation method is presented in this section.
Domain distance and domain density calculation methods are presented based on the KNN method \cite{ex89, ex95}.
These methods are very useful and handy on large-scale datasets that likely contain varying distribution densities in actual applications.

To more precisely explain the locality of VDD data, we propose a new definition of domain density based on the KNN method.
Given a dataset $X$, the $KNN$-distance and $KNN$-density of each data point in $X$ are calculated, respectively.

\textbf{Definition 3.2}
\textit{$KNN$-Distance. Given a dataset $X$, the $KNN$-distance of each data point $x_{i}$ refers to the average distance of $x_{i}$ to its $k$ nearest neighbors.
The $KNN$-distance of each data point $x_{i}$ is defined as $KDist_{i}$:}
\begin{equation}
\label{eq03}
KDist_{i} = \frac{1}{K} \sum_{j \in N(x_{i})}{d_{ij}},
\end{equation}
where $K$ is the number of neighbors of $x_{i}$ and $N(x_{i})$ is the set of its neighbors.
Based on the $KNN$-distance, we calculate the $KNN$-density for each data point.

\textbf{Definition 3.3}
\textit{$KNN$-Density. The $KNN$-density of the data point $x_{i}$ in dataset $X$ refers to the reciprocal of the $KNN$-distance.
The smaller $KNN$-density of a data point, indicating that this data point is located in a more sparse area.
The $KNN$-density of data point $x_{i}$ is defined as $KDen_{i}$:}
\begin{equation}
\label{eq04}
\begin{aligned}
KDen_{i} &= \frac{1}{KDist_{i}}= \frac{K}{\sum\limits_{j \in N(x_{i})}{d_{ij}}}.
\end{aligned}
\end{equation}

After obtaining the $KNN$-distance and $KNN$-density, the domain-adaptive density of each data point is defined.
We treat the set of each data point $x_{i}$ and its neighbors $N(x_{i})$ as a subgroup to observe its density distribution in $X$.

\textbf{Definition 3.4}
\textit{Domain Density. The domain density of each data point $x_{i}$ in dataset $X$ is the sum of the $KNN$-density of $x_{i}$ and the weighted $KNN$-density of its $K$-nearest neighbors.
The domain density of the data point $x_{i}$ is defined as $\partial_{i}$:}
\begin{equation}
\label{eq05}
\partial_{i} =KDen_{i} + \sum_{j \in N(x_{i})}{\left( KDen_{j} \times w_{j}\right)},
\end{equation}
where $w_{j} = \frac{1}{d_{ij}}$ is the weighted value of the $KNN$-density between each neighbor $x_{j}$ and $x_{i}$.
Compared to the $KNN$-density, the domain density can better reflect the density distribution of data points in the local area.

\begin{figure}[!ht]
  \centering
  \includegraphics[width=3.3in]{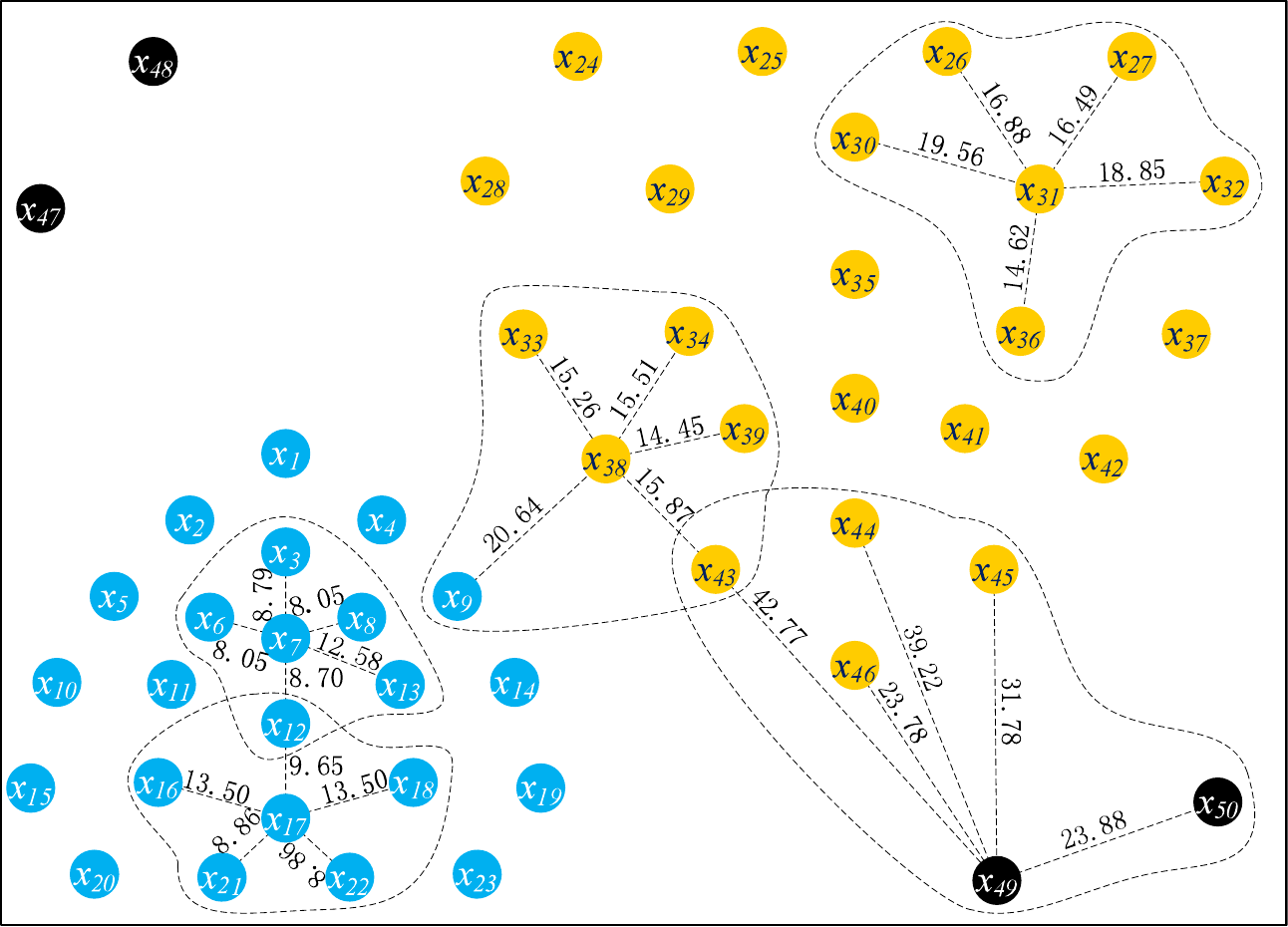}
  \caption{Example of domain density calculation (partial).}
  \label{img02}
\end{figure}

Given a $2$-dimensional dataset with 50 samples as an example, we calculate the distances among data points, as shown in Fig. \ref{img02}.
Set $K=5$, for each point, $KNN$ neighbors of $x_{7}$ are $x_{6}$, $x_{8}$, $x_{12}$, $x_{3}$, and $x_{13}$, with the distances of 8.05, 8.05, 8.70, 8.79, and 12.58, respectively.
According to Eq. (\ref{eq03}), it is easy to obtain the $KNN$-distance of $x_{7}$ is equal to 9.23.
Hence, the $KNN$-density $KDen_{7}$ of $x_{7}$ is equal to 0.11 and the domain density $\partial_{7}$ of $x_{7}$ is 0.16.
In the same way, $KNN$-distances and $KNN$-densities of the neighbors of $x_{7}$ are calculated successively.
We further calculate the domain density of these data points: $\partial_{6}=\partial_{8}=0.15$, $\partial_{3}=0.12$, $\partial_{12}=0.09$, and $\partial_{13}=0.12$.
It is obvious that $x_{7}$ has a higher value of the domain density than that of its neighbors, reaching the value at 0.16.

\subsection{Clustering Decision Parameter Measurement}
\begin{figure}[!ht]
  \centering
  \includegraphics[width=3.3in]{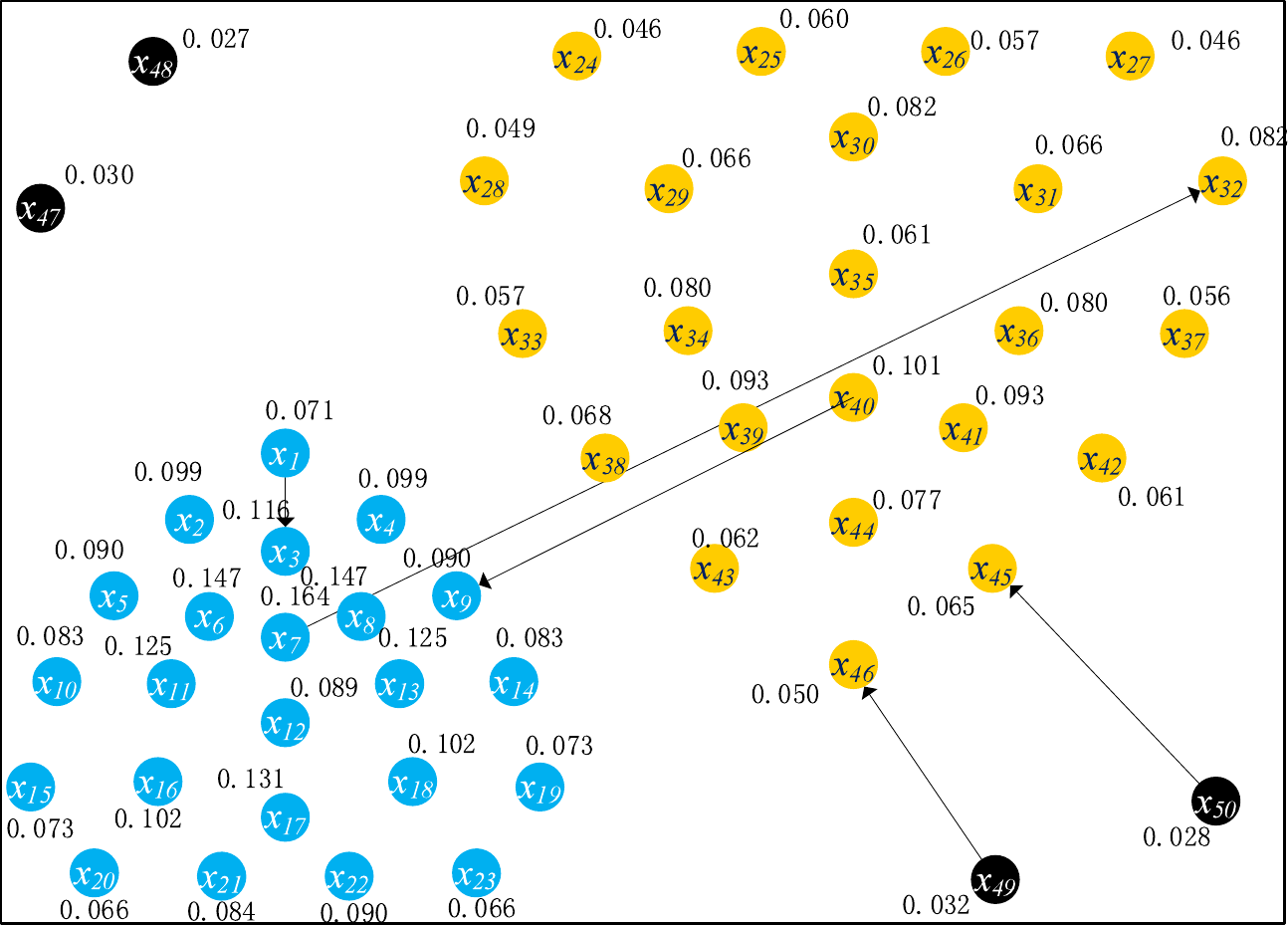}
  \caption{Example of Delta distance calculation (partial).}
  \label{img03}
\end{figure}

Based on the domain density, the Delta distance of each data point is computed as a clustering decision parameter.
As defined in Eq. (\ref{eq02}), Delta distance $\delta_{i}$ of $x_{i}$ is measured by calculating the shortest distance between $x_{i}$ and any other points with higher densities.
In such a case, only the points with the highest global density have the maximum value of the Delta distance.
The domain density peak in a sparse region yields a Delta distance value that is lower than the remaining points in a relatively dense region.
An example of Delta distances of a dataset is shown in Fig. \ref{img03}.

In Fig. \ref{img03}, domain densities of all data points are calculated.
Since $x_{7}$ owns the highest domain density, Delta distance $\delta_{7}$ is the distance between $x_{7}$ and the point $x_{32}$ farthest against it.
Namely, $\delta_{7} = d_{(7,32)}=103.92$.
Because the remaining data points do not have the highest domain density, the Delta distances of them are the shortest distance between them and the point with higher density.
For example, $\delta_{1}= d_{(1,2)}=7.52$,  $\delta_{40}= d_{(40,9)}=42.76$,  $\delta_{49}= d_{(49,46)}=23.81$, and $\delta_{50}= d_{(50,45)}=30.44$.

Considering that a dataset has multiple regions with different densities, domain densities of data points in a dense region are higher than that of points in a sparse region.
To adaptively identify the density peaks of each region, we update the definition of domain-adaptive density by combining the values of the domain density and Delta distance.
The domain-adaptive density $\partial_{i}$ of each data point $x_{i}$ is updated as:
\begin{equation}
\label{eq081}
\partial_{i} =\partial_{i} \times  \delta_{i}=
\begin{cases}
\partial_{i} \times \max\limits_{j}{(d_{ij})}, & \text{ if } \partial_{i} = \partial_{max};\\
\partial_{i} \times \min\limits_{j:\partial_{j} > \partial_{i}}{(d_{ij})}, & \text{ otherwise }.
\end{cases}
\end{equation}

There are three levels of domain density for data points: global density maximum, domain density maximum, and normal density.
(1) It is easy to identify the point with the highest global density and set it as a cluster center.
     For the global density maximum point, we set the largest distance between this point and any other point as its Delta distance.
(2) For a density maximum point $x_{i}$ of a region, the point $x_{j}|\min_{j:\partial_{j} > \partial_{i}}(d_{ij})$ must be in another region with a greater density rather than in the current region.
     Therefore, to clearly identify the density peaks of a region, we multiply the domain density and Delta distance for each point.
(3) For the remaining points in each region, both of their domain densities and Delta distances are much smaller than that of the peak points of the same region.

Based on the values of domain density $\partial$ and Delta distance $\delta$, a clustering decision graph is drawn to identify the candidate cluster centers.
In the clustering decision graph, the horizontal axis represents $\partial$ and the vertical axis represents $\delta$.
Points having high values of $\partial$ and $\delta$ are considered as cluster centers, while points with a low $\partial$ and a high $\delta$ are considered as outliers.

The process of domain density and Delta distance calculation of DADC is presented in Algorithm \ref{alg01}.
Assuming that the number of data points in $X$ is equal to $n$, for each data point $x_{i}$ in $X$, we calculate its K-nearest neighbors and obtain its domain density.
Therefore, the computational complexity of Algorithm \ref{alg01} is $O(n)$.

\begin{algorithm}[!ht]
\caption{Domain-adaptive density and Delta distance calculation of DADC.}
\label{alg01}
\begin{algorithmic}[1]
\REQUIRE ~\\
    $X$: the dataset for clustering;\\
    $K$: the number of neighbors of each data point;\\
\ENSURE ~\\
   ($\partial$, $\delta$): The domain-adaptive densities and Delta distances of the data points of $X$.\\
\STATE calculate distance matrix $D$ for $X$;
\FOR {each $x_{i}$ in $X$}
\STATE obtain $K$-nearest neighbors $N(x_{i})$ of $x_{i}$;
\STATE calculate $KNN$-distance $KDist_{i} \leftarrow \frac{1}{K} \sum\limits_{j \in N(x_{i})}{d_{ij}}$;
\STATE calculate $KNN$-density $KDen_{i} \leftarrow \frac{K}{\sum\limits_{j \in N(x_{i})}{d_{ij}}}$;
\STATE calculate domain density $\partial_{i} \leftarrow KDen_{i} + \sum\limits_{j \in N(x_{i})}{\left(KDen_{j} \times \frac{1}{d_{ij}}\right)}$;
\ENDFOR
\STATE get the maximum domain density $\partial_{max}$ $\leftarrow$ max($\partial$);
\FOR {each $x_{i}$ in $X$}
\IF {$\partial_{i}$ == $\partial_{max}$}
\STATE set Delta distance $\delta_{i} \leftarrow$ max($d_{ij}$);
\ELSE
\STATE set Delta distance $\delta_{i} \leftarrow \min\limits_{j:\partial_{j} > \partial_{i}}{d_{ij}}$;
\ENDIF
\STATE calculate domain-adaptive density $\partial_{i} \leftarrow \partial_{i} \times \delta_{i}$;
\ENDFOR
\RETURN ($\partial$, $\delta$).
\end{algorithmic}
\end{algorithm}

\section{Cluster Self-identification Method}
\label{section4}
For data with ED or MDDM features, a large number of density peaks with similar values can be identified, which results in cluster fragmentation.
In this section, aiming at the problem of cluster fragmentation, we propose a cluster self-identification method to extract initial cluster centers by automatically determining the parameter thresholds of the clustering decision graph.
Then, a Cluster Fusion Degree (CFD) model is proposed to evaluate the relationship of adjacent clusters.
Finally, a CFD-based cluster self-ensemble method is proposed to merge the fragmented clusters.

\subsection{Problem Definitions}
\label{section4.1}
Based on domain-adaptive densities and Delta distances, candidate cluster centers of a dataset can be obtained from the corresponding clustering decision graph.
After the cluster centers are identified, each of the remaining data points is assigned to the cluster to which the nearest and higher-density neighbors belong.

(1) Decision-parameter threshold determination.

\begin{figure}[!ht]
  \centering
  \subfigure[Frame dataset]{\includegraphics[width=1.6in]{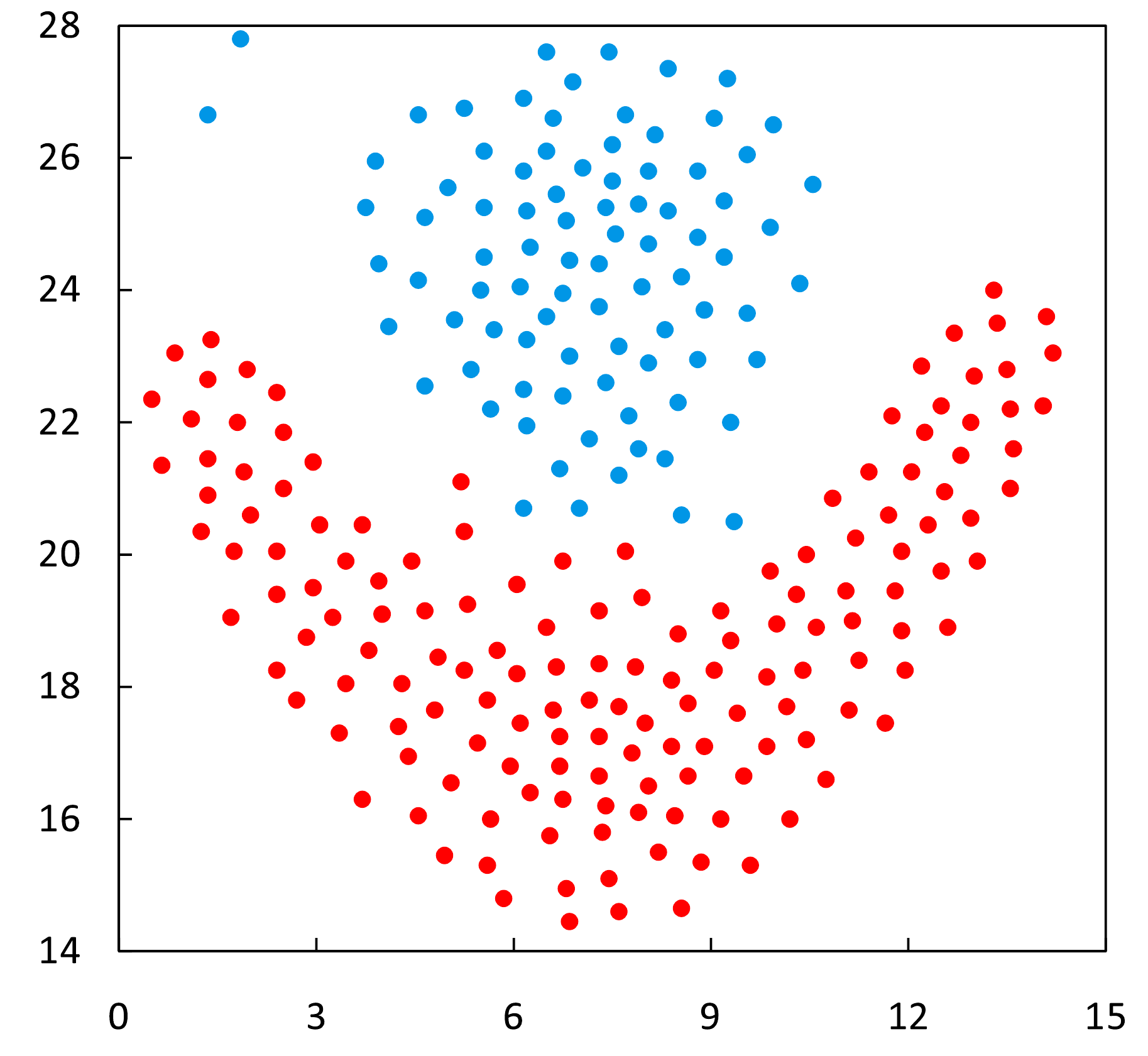}}
  \subfigure[Decision graph for Frame]{\includegraphics[width=1.6in]{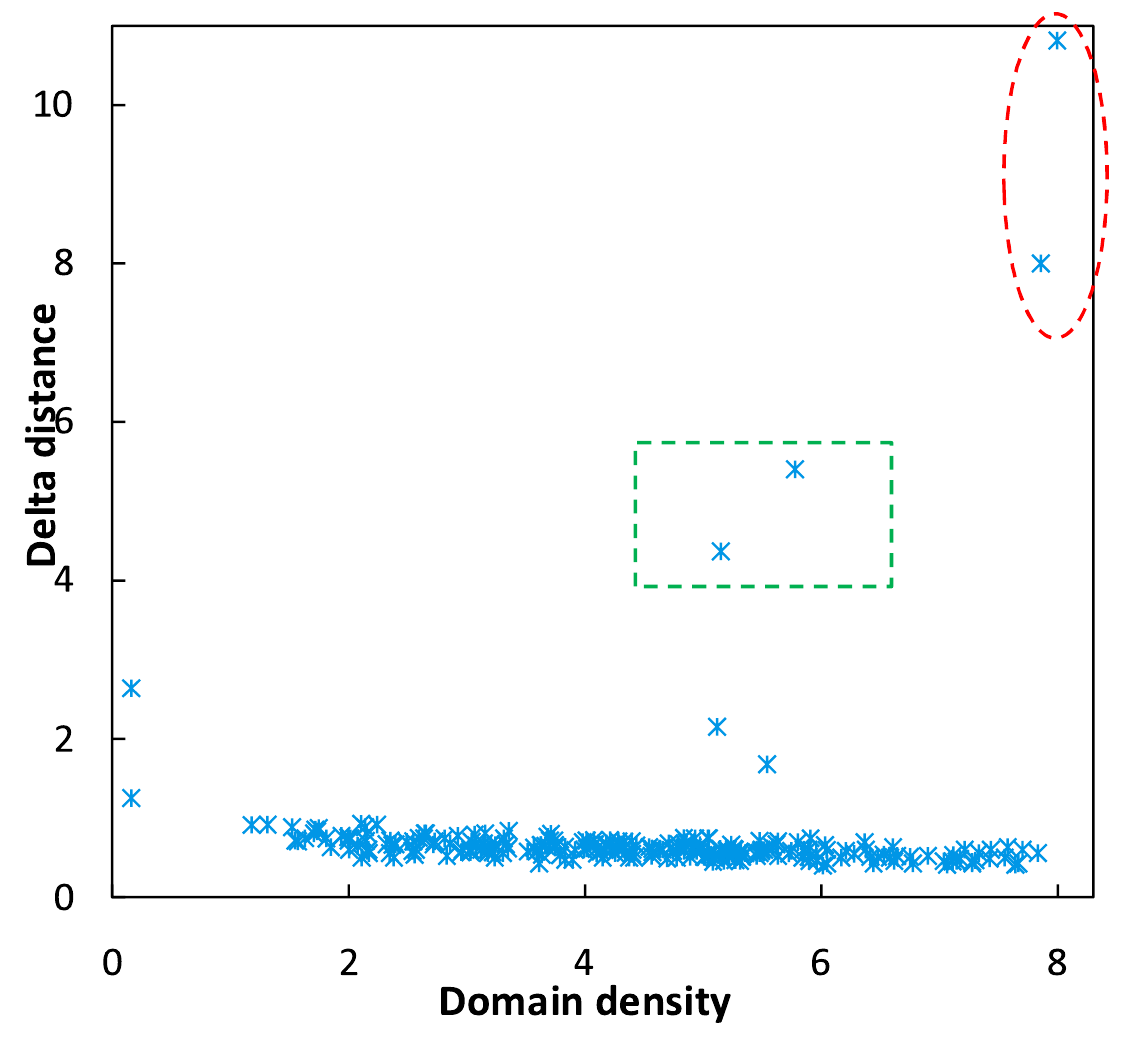}}
  \caption{Decision-parameter threshold determination.}
  \label{img07}
\end{figure}

A limitation of the CFSFDP algorithm is that how to determine the thresholds of the decision parameters in the clustering decision graph.
In CFSFDP, data points with high values of both local density and Delta distance are regarded as cluster centers.
But in practice, these parameter thresholds are often set manually.
An example of the \emph{Frame} dataset and the corresponding clustering decision graph are shown in Fig. \ref{img07}.

Fig. \ref{img07} (b) is the clustering decision graph of CFSFDP for the dataset in Fig. \ref{img07} (a).
It is difficult to make a decision whether only the points in the red box or those in both red and blue boxes should be regarded as cluster centers.
Therefore, how to determine the threshold values of decision parameters in an effective way is an important issue of our algorithm.

(2) Cluster fragmentation on MDDM or ED Data.

Most DPC algorithms have limitations of cluster fragmentation on the datasets with multiple domain-density maximums (MDDM) or equilibrium distribution (ED).

\textbf{Definition 4.1}
\textit{MDDM Dataset.
Given a dataset, the domain-adaptive densities of data points in the dataset are calculated.
If multiple points with the same highest domain density coexist in a region, we call the dataset holds the characteristic of multiple domain-density maximums.
A dataset with multiple domain-density maximums is defined as an MDDM dataset.}

\textbf{Definition 4.2}
\textit{ED Dataset.
Given a dataset, the domain densities of data points in the dataset are calculated.
If each data point has the same value of domain density, the dataset is under an equilibrium distribution and is defined as an ED dataset.
In such a case, each data point having the same value of domain density is regarded as a domain density peak and further considered as a candidate cluster center.}

\begin{figure}[!ht]
  \centering
  \subfigure[MDDM dataset]{\includegraphics[width=1.6in]{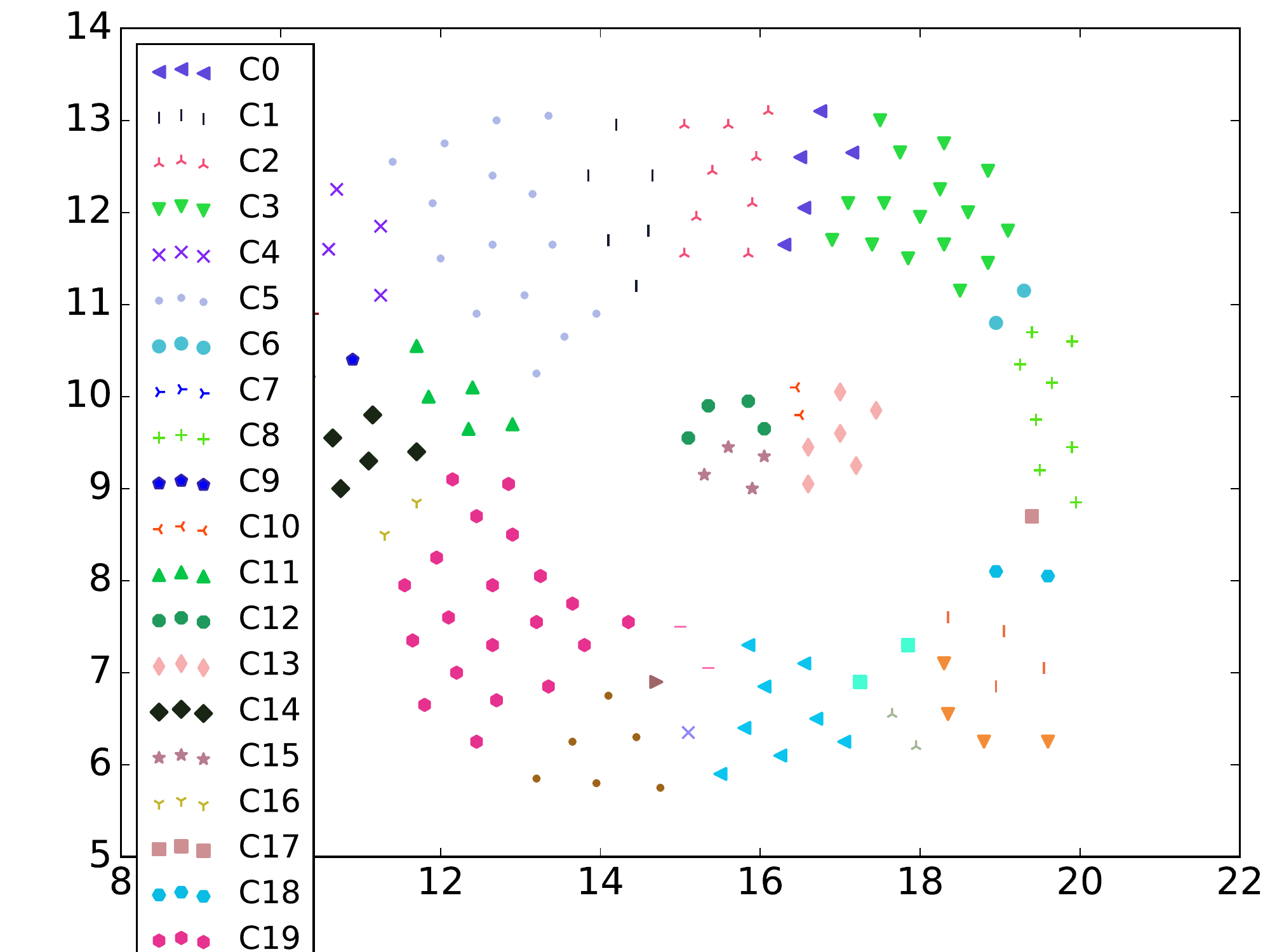}}
  \subfigure[Decision graph]{\includegraphics[width=1.6in]{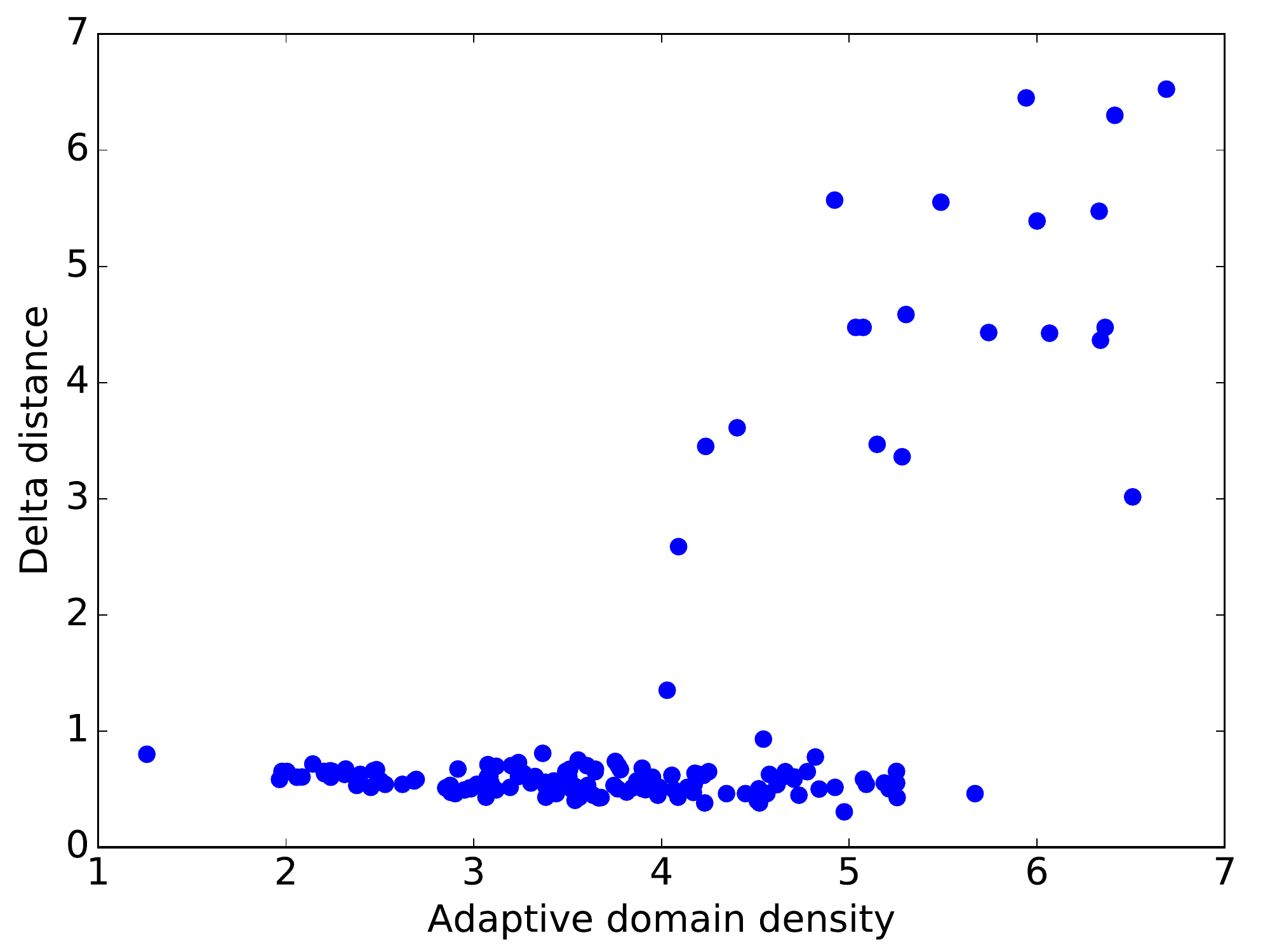}}
  \subfigure[ED dataset]{\includegraphics[width=1.6in]{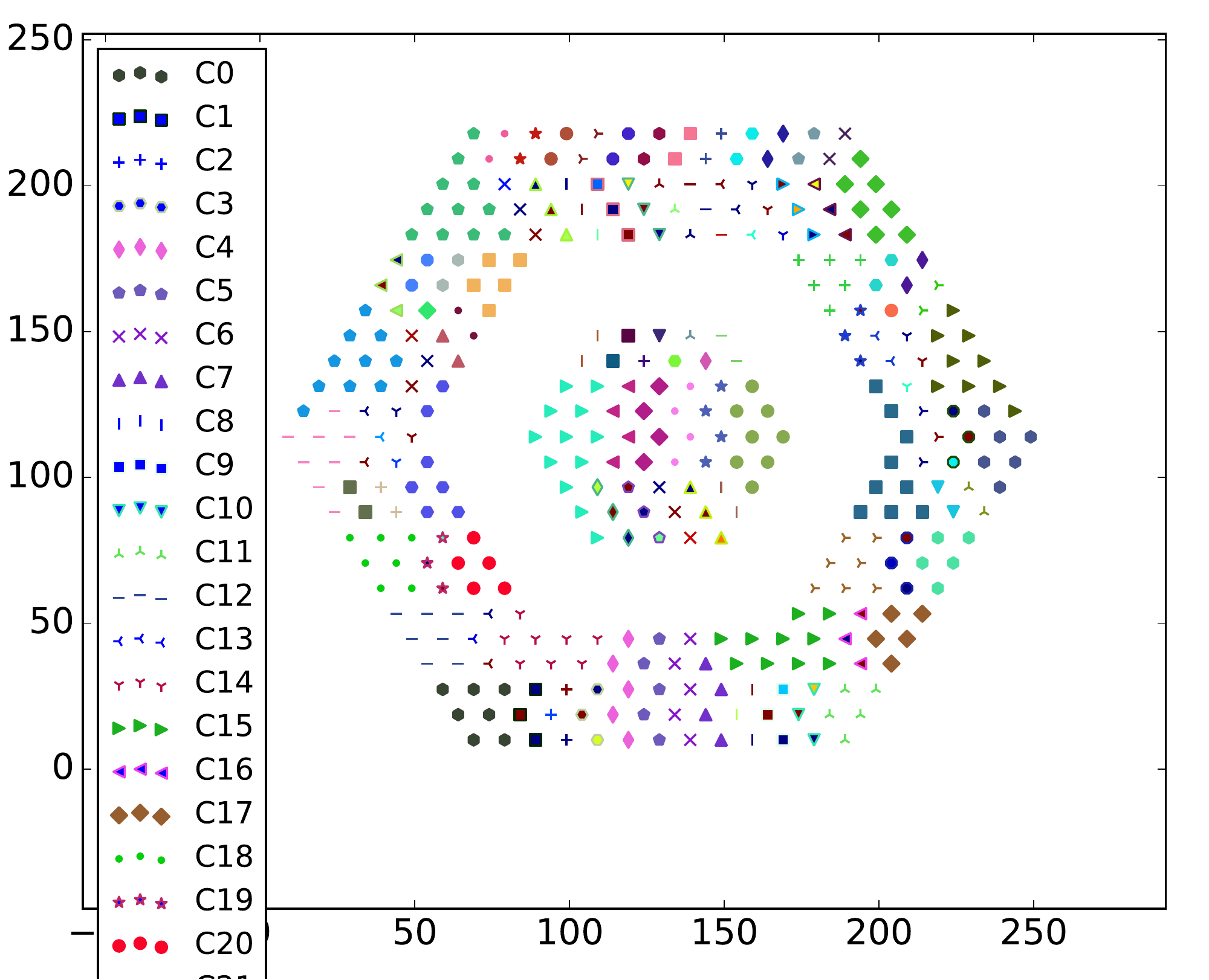}}
  \subfigure[Decision graph]{\includegraphics[width=1.6in]{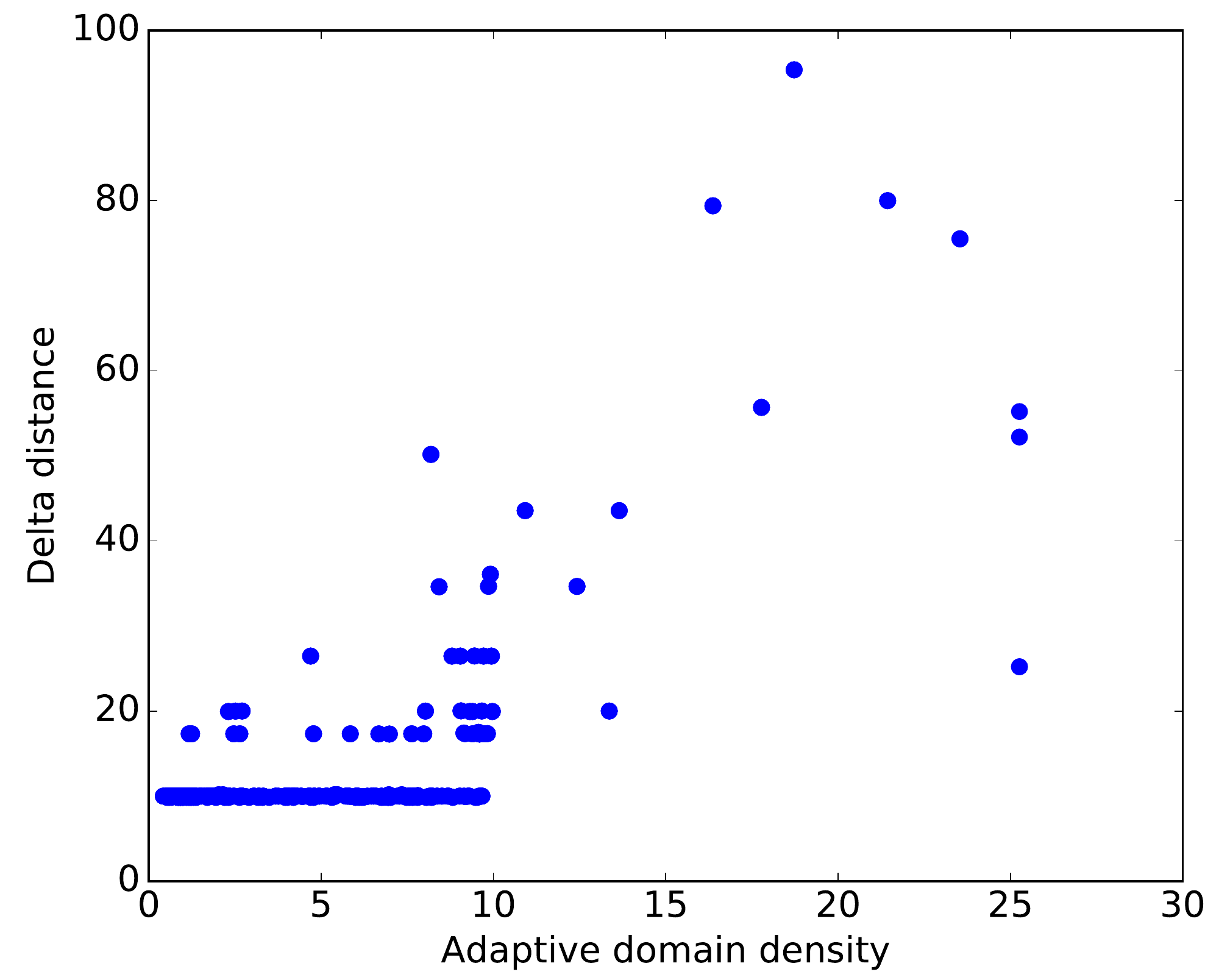}}
  \caption{Clustering decision graph of MDDM and ED dataset.}
  \label{img04}
\end{figure}

Clustering results of the CFSFDP algorithm depend on a strict constraint that only one local density maximum is assumed to exist in each candidate cluster.
However, when there exist multiple local density maximums in a natural cluster, CFSFDP might lead to the problem of cluster fragmentation.
Namely, a cluster is split into many fragmented clusters.
Two examples of the clustering decision graph of CFSFDP on MDDM and ED datasets are shown in Fig. \ref{img04}.

In Fig. \ref{img04} (b), there are as many as 29 decision points that hold high values of both domain density $\partial$ and Delta distance $\delta$.
In such a case, the dataset is divided into 29 fragmented clusters instead of 2 natural clusters as shown in Fig. \ref{img04} (a).
As shown in Fig. \ref{img04} (c), there are two isolated regions in the dataset, and data points in each region are equilibration distributed.
Hence, this dataset is expected to be divided into 2 natural clusters.
However, lots of points exhibit similar values of local/domain densities and are regarded as cluster centers, as shown in Fig. \ref{img04} (d).
Consequently, this dataset is incorrectly divided into numerous fragmented sub-clusters rather than the expected two clusters.

\subsection{Initial Cluster Self-identification}
\label{section4.2}

\subsubsection{Cluster Center Identification}
\begin{figure}[!ht]
  \centering
  \includegraphics[width=2.0in]{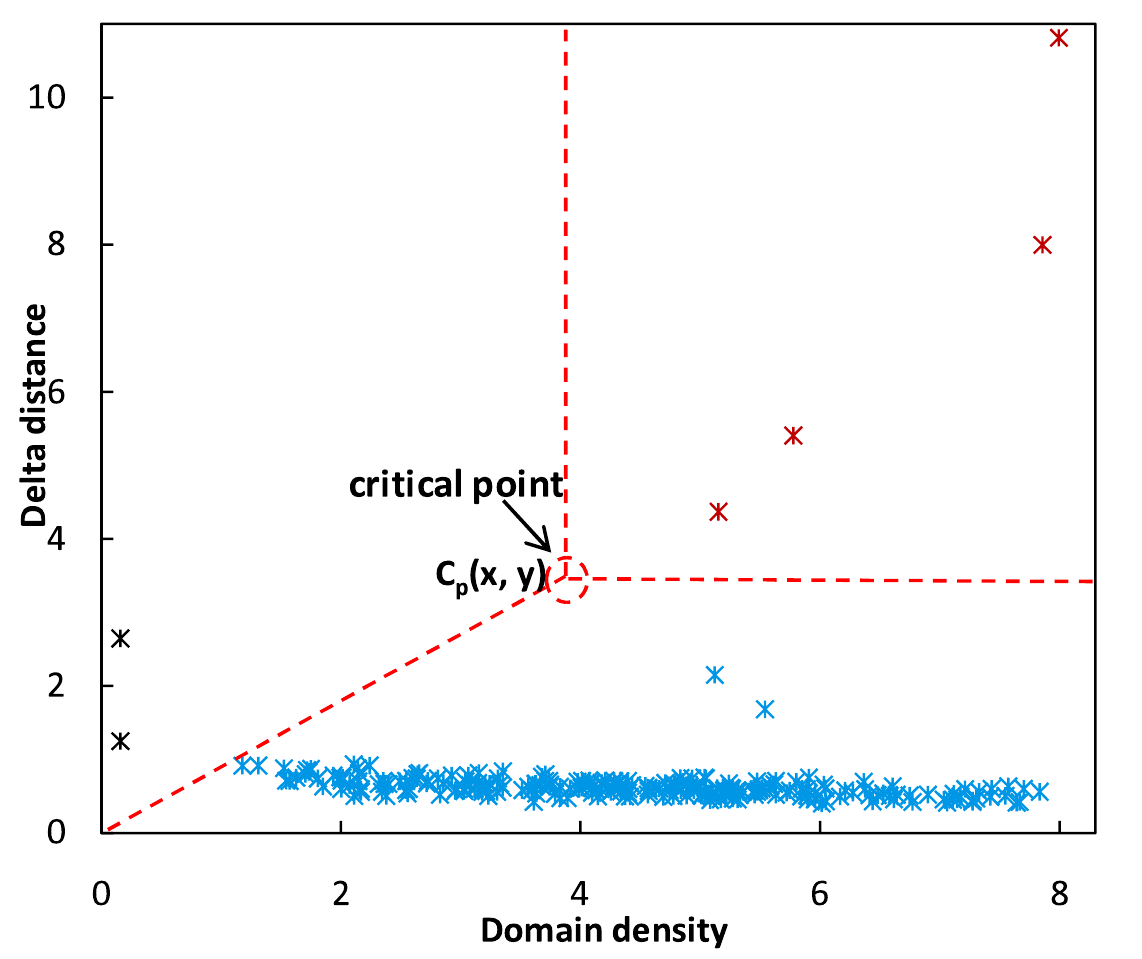}
  \caption{Critical point of a clustering decision graph.}
  \label{img08}
\end{figure}

We propose a self-identification method to automatically extract the cluster centers based on the clustering decision graph.
To automatically determine the parameters threshold values of domain density and Delta distance, a critical point of the clustering decision graph is defined.
The critical point $C_{p}(x,~y)$ of a clustering decision graph is defined as a splitting point by which the candidate cluster centers, outliers, and remaining points can be divided obviously, as shown in Fig. \ref{img08}.

As the assumption of the CFSFDP and DADC algorithms, cluster centers are the points with relatively domain-density peaks, while outliers have the lowest domain densities.
It is easy to get a conclusion that the values of domain density of density peaks are obviously different against to that of outliers.
Therefore, we take the middle value of the maximum domain density as the horizontal axis value of the critical point.
Namely, $C_{p}(x) = \frac{\partial_{max}}{2}$.
In addition, based on extensive experiments and applications, it is a effectiveness solution to set the vertical axis value of the critical point as one quarter of the maximum value of the Delta distance.
Namely, $C_{p}(y) = \frac{\delta_{max}}{4}$.
Therefore, the value of critical point $C_{p}(x,~y)$ of the clustering decision graph is defined as:
\begin{equation}
\label{eq09}
C_{p}(x, y) = \left(\frac{\partial_{max}}{2}, \frac{\delta_{max}}{4}\right),
\end{equation}
where $\delta_{max}$ and $\partial_{max}$ are the maximum values of $\delta$ and $\partial$.

Based on the critical point, data points in the clustering decision graph can be divided into three subsets, namely, cluster centers, outliers, and remaining points.
The division method of data points in the clustering decision graph is defined as:

\begin{equation}
\label{eq10}
\Lambda(x_{i}) =
\begin{cases}
\begin{array}{lcl}
\text{cluster centers}, & \partial_{i} > C_{p}(x),~ \delta_{i} > C_{p}(y); \\
\text{outliers},    & \partial_{i} < C_{p}(x),~ \delta_{i} > \frac{C_{p}(y) \times \partial_{i}}{C_{p}(x)} ;\\
\text{remaining points},& \text{otherwise},\\
\end{array}
\end{cases}
\end{equation}
where $\Lambda(x_{i})$ refers to the subset $x_{i}$ belong to.
In Fig. \ref{img08}, after getting the value of the critical point, data points in the decision graph can be easily divided into three subsets.
Red points are detected as candidate cluster centers.
Black points have low values of domain density and high values of Delta distance, identifying as outliers and removed from the clustering results.
Blue points refer to the remaining data points, which are assigned to the related clusters in the next step.
Hence, initial cluster centers of the dataset are obtained with few parameter requirements and minimum artificial intervention.

\subsubsection{Remaining Data Point Assignment}
\begin{figure}[!ht]
  \centering
  \includegraphics[width=2.0in]{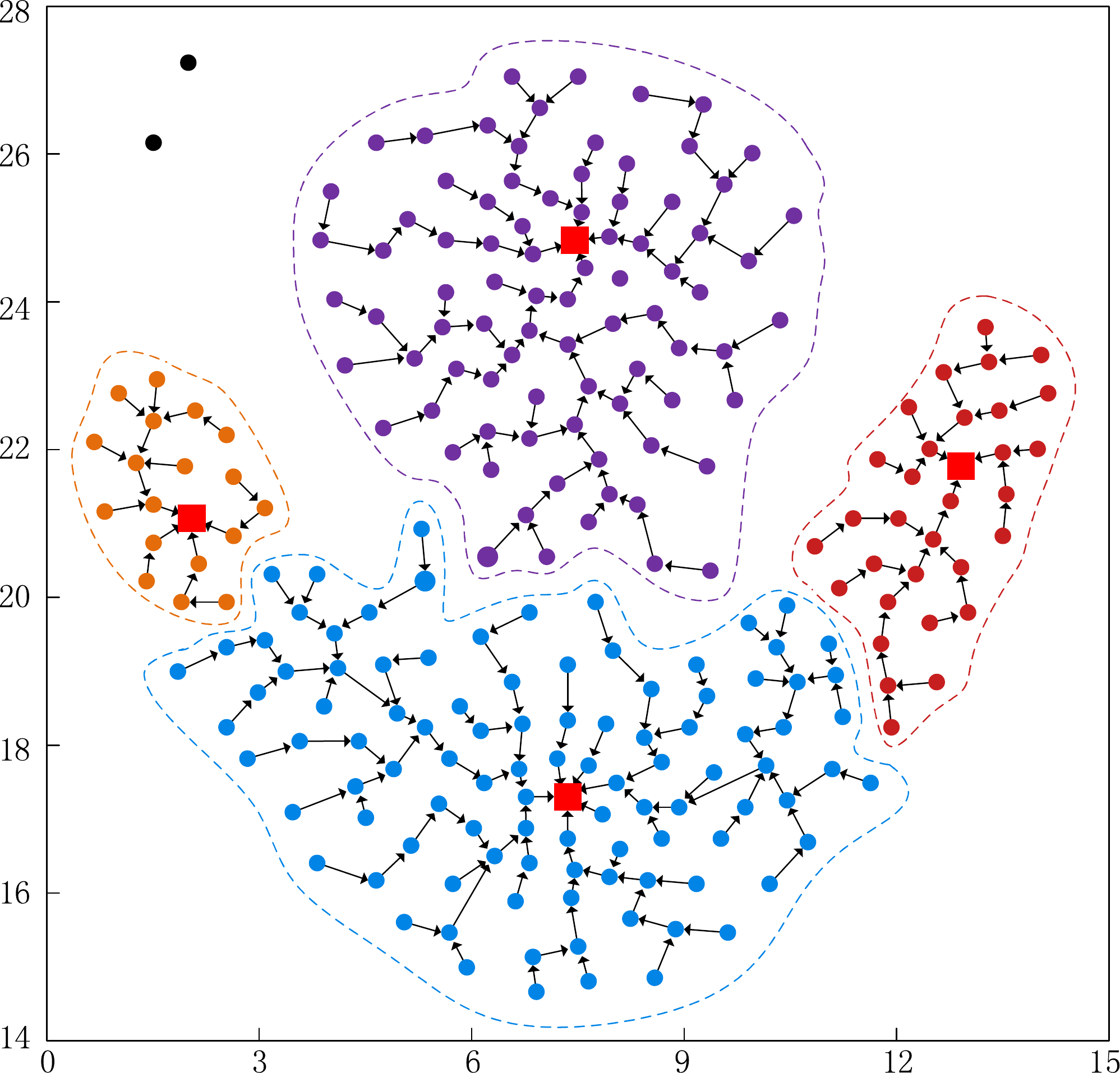}
  \caption{Example of the remaining data points assignment.}
  \label{img09}
\end{figure}

After cluster centers being detected, each of the remaining data points is assigned to the cluster that the nearest and higher domain-density neighbors belong to.
For each remaining data point $x_{i}$, the neighbors with a higher density are labeled as $N^{'}(x_{i})$.
For a data point $x_{j} \in N^{'}(x_{i})$ with the shortest distance $d_{ij}$, if $x_{j}$ has been assigned to a cluster $c_{a}$, then, $x_{i}$ is also assigned to $c_{a}$.
Otherwise, the cluster of $x_{j}$ is further measured iteratively.
Repeat this step, until all of the remaining data points are assigned to the related clusters.
An example of the remaining data points assignment is shown in Fig. \ref{img09}.
The process of initial cluster self-identification of the DADC algorithm is presented in Algorithm \ref{alg02}.

\begin{algorithm}[!ht]
\caption{Cluster center self-identification of the DADC algorithm.}
\label{alg02}
\begin{algorithmic}[1]
\REQUIRE ~\\
    $X$: The raw dataset for clustering;\\
    $\partial$: The domain densities of the data points of $X$;\\
    $\delta$: The Delta distances of the data points of $X$;\\
\ENSURE ~\\
    $IC_{X}$: The initial clusters of $X$.\\
\STATE get the maximum domain density $\partial_{max}$ $\leftarrow$ max($\partial$);
\STATE get the maximum Delta distance $\delta_{max}$ $\leftarrow$ max($\delta$);
\STATE calculate the critical point $C_{p}(x, y) = \left(\frac{\partial_{max}}{2}, \frac{\delta_{max}}{4}\right)$;
\FOR {each $x_{i}$ in $X$}
\IF  {$\partial_{i} > C_{p}(x)$ and $\delta_{i} > C_{p}(y)$}
\STATE append to the set of cluster centers $\Lambda_{c}$ $\leftarrow$ $x_{i}$;
\ELSIF {$\partial_{i} < C_{p}(x)$ and $\delta_{i} > \frac{C_{p}(y) \times \partial_{i}}{C_{p}(x)}$}
\STATE append to the set of outliers $\Lambda_{o}$ $\leftarrow$ $x_{i}$;
\ELSE
\STATE append to the set of remaining data points $\Lambda_{n}$ $\leftarrow$ $x_{i}$;
\ENDIF
\ENDFOR
\FOR {each $x_{i}$ in $\Lambda_{n}$}
\STATE append $x_{i}$ to the nearest cluster $IC_{X}$;
\ENDFOR
\RETURN $IC_{X}$.
\end{algorithmic}
\end{algorithm}

Algorithm \ref{alg02} consists of two steps: cluster center identification and remaining data point assignment.
Assuming that the number of data points in $X$ is $n$, the number of cluster centers is $m$, and that of remaining data points is $n'$.
In general, the number of cluster centers and outliers is far less than the remaining points, which shows that $m + n' \approx n$. Therefore, the computational complexity of Algorithm \ref{alg02} is $O(n)$.

Depending on the cluster self-identification method of DADC, we can obtain cluster centers and initial clusters quickly and simply.
Despite the number of cluster centers might be more than the real ones in this way and caused many fragmented clusters, it does not lead to the clustering result that multiple clusters are wrongly classified as a cluster.
Focus on the scenario of fragmented clusters, we introduce a cluster self-ensemble method to merge the preliminary clustering results.

\subsection{Fragmented Cluster Self-ensemble}
\label{section4.3}
To address the limitation of cluster fragmentation of DADC on MDDM datasets, a fragmented cluster self-ensemble method is proposed in this section.
As we all know, the basic principle of clustering analysis is that individuals in the same cluster have high similarities with each other, while different from individuals in different clusters.
Therefore, to find out which clusters are misclassified into multiple subclusters, we propose an inter-cluster similarity measurement and cluster fusion degree model for fragmented cluster self-ensemble.
Clusters with a superior density similarity and cluster fusion are merged into the same cluster.

\textbf{Definition 4.3}
\textit{Inter-cluster Density Similarity (IDS).
Inter-cluster density similarity between two clusters refers to the degree of similarity degree of their cluster densities.
The average density of a cluster is the average value of the domain densities of all data points in the cluster.}

Let $\mathcal{S}_{a,b}$ be the inter-cluster density similarity between cluster $c_{a}$ and $c_{b}$.
The larger the value of $\mathcal{S}_{a,b}$, the more similar is the density of the two clusters.
$\mathcal{S}_{a,b}$ is defined as:
\begin{equation}
\label{eq11}
\begin{aligned}
\mathcal{S}_{a,b} &= \frac{2\sqrt{\overline{KDen_{c_{a}}} \times \overline{KDen_{c_{b}}}}}{\overline{KDen_{c_{a}}} +\overline{KDen_{c_{b}}}},
\end{aligned}
\end{equation}
where $\overline{KDen_{c_{a}}} = \frac{1}{|c_{a}|}\sum_{i \in c_{a}}{KDen_{i}}$ and $0 < \mathcal{S}_{a,b} = \mathcal{S}_{b,a} \leq 1$.
Let $f(\frac{u}{v})=\frac{2\sqrt{uv}}{u+v}=\frac{2}{\sqrt{\frac{u}{v}}+\sqrt{\frac{v}{u}}}$, where $\frac{u}{v}\in (0,1]$.
Let $x=\frac{u}{v}$, then $f(x)=\frac{2x}{x^2+1}$.
Since $f'(x)=\frac{2(1-x^2)}{(x^2+1)^2}\geq 0$, then $f(x)$ is a strictly monotonically increasing function and $f_{max}(x)=f(1)=1$.
Hence the closer the value of $x$ is to 1, the more similar are the two clusters $c_{a}$ and $c_{b}$.

In addition, the distance between every two clusters is considered.
In the relevant studies, different methods were introduced to calculate the distance between two clusters, such as the distance between the center points, the nearest points, or the farthest points of the two clusters \cite{ex09}.
However, these measures are easily affected by noise or outliers, while noisy data elimination is another challenge.
We propose an innovative method to measure the distance between clusters.
Crossing points between every two clusters are found and the crossover degree of the clusters is calculated.

For each boundary point $x_{i}$ in cluster $c_{a}$, let $N(x_{i})$ be the $K$-nearest neighbors of $x_{i}$.
We denote $N_{(i,b)}$ as a set of points in $N(x_{i})$ belonging to cluster $c_{b}$, and $N_{(i,a)}$ be a set of points in $N(x_{i})$ belonging to $c_{a}$.
If the amount of neighbors belonging to $c_{b}$ is close to that of neighbors belonging to the current cluster $c_{a}$, then, $x_{i}$ is defined as a crossing point of $c_{a}$, and is represented as $x_{i} \in CP_{(a \rightarrow b)}$.
The crossover degree $\emph{c}_{(i, ~a \rightarrow b)}$ of a crossing point $x_{i}$ in $c_{a}$ between clusters $c_{a}$ and $c_{b}$ is defined as:

\begin{equation}
\label{eq12}
\begin{aligned}
\emph{c}_{(i, a \rightarrow b)} = \frac{2\sqrt{|N_{(i,a)}| \times |N_{(i,b)}|}}{|N_{(i,a)}| +|N_{(i,b)}|},
\end{aligned}
\end{equation}
where  $N_{(i,a)} \in (N(x_{i}) \bigcap c_{a})$, $N_{(i,b)} \in (N(x_{i}) \bigcap c_{b})$, and $0 < \emph{c}_{(i, a \rightarrow b)} \leq 1$.
Based on crossover degrees of all crossing points of each cluster, we can define the crossover degree between every two clusters.

\textbf{Definition 4.4}
\textit{Cluster Crossover Degree (CCD).
Cluster crossover degree $\mathcal{C}_{a,b}$ of two clusters $c_{a}$ and $c_{b}$ is calculated by the sum of the crossover degrees of all crossing points between $c_{a}$ and $c_{b}$. The formula of CCD is defined as:}
\begin{equation}
\label{eq13}
\begin{aligned}
\mathcal{C}_{a,b} = \sum_{x_{i} \in CP_{(a \rightarrow b)}}{\emph{c}_{(i, a \rightarrow b)}} +
\sum_{x_{j} \in CP_{(b \rightarrow a)}}{\emph{c}_{(j, b \rightarrow a)}}.
\end{aligned}
\end{equation}

To measure whether the data points in a cluster have similar domain densities, we give a definition of the cluster density stability.
By analyzing the internal density stability of the clusters to be merged and that of the merged cluster, we can determine whether the merger is conducive to the stability of these clusters.
The internal density stability of clusters is an important indicator of cluster quality.

\textbf{Definition 4.5}
\textit{Cluster Density Stability (CDS).
Cluster density stability is the reciprocal of the cluster density variance, which is calculated by the deviation between the domain density of each point and the average domain density of the cluster.
The larger the CDS of a cluster, the smaller domain density differences of each point in the cluster.
The CDS of a cluster $c_{a}$ is defined as:}
\begin{equation}
\label{eq14}
\begin{aligned}
\emph{d}_{a} =\log{\left(\sqrt{\sum\limits_{i \in c_{a}}{(KDen_{i} - \overline{KDen_{c_{a}}})^{2}}}\right)},
\end{aligned}
\end{equation}
where $\overline{KDen_{c_{a}}}$ is the average value of domain densities of the data points in $c_{a}$, and $|c_{a}|$ is the number of data points in $c_{a}$.
A cluster with a high CDS means that data points in the cluster have low domain-density differences.
Namely, most data points in the same cluster have similar domain densities.

For two clusters $c_{a}$ and $c_{b}$ with high inter-cluster density similarity and high crossing degree, we can further calculate their CDS.
Assuming that $\emph{d}_{a}$ and $\emph{d}_{b}$ are the CDSs of $c_{a}$ and $c_{b}$, and $\emph{d}_{a+b}$ is that of the new cluster merged from $c_{a}$ and $c_{b}$.
CDS $\mathcal{D}_{a,b}$ between $c_{a}$ and $c_{b}$ is calculated in Eq. (\ref{eq15}):
\begin{equation}
\label{eq15}
\mathcal{D}_{a,b} =\frac{\emph{d}_{a+b}}{\emph{d}_{a}}\times \frac{\emph{d}_{a+b}}{\emph{d}_{b}}.
\end{equation}
If the CDS of the merged cluster is close to the average value of $\emph{d}_{a}$ and $\emph{d}_{b}$, it indicates that the merger of the two clusters does not reduce their overall density stability.

Based on the above indicators of clusters, including inter-cluster density similarity, cluster crossover degree, and cluster density stability, the definition of cluster fusion degree is proposed.

\textbf{Definition 4.6}
\textit{Cluster Fusion Degree (CFD).
Cluster fusion degree of two clusters is the degree of the correlation between the clusters in terms of the location and density distribution, which is calculated depending upon the values of IDS, CCD, and CDS.
Two clusters with a high degree of fusion should satisfy the following conditions:
(1) having a high value of IDS,
(2) having a high value of CCD,
and (3) the CDS of the merged cluster should be close to the average value of the two initial clusters' CDSs.
If two adjacent and crossed clusters hold a high IDS and similar CDS, they have a high fusion degree.}

\begin{figure}[!ht]
  \centering
  \includegraphics[width=2.5in]{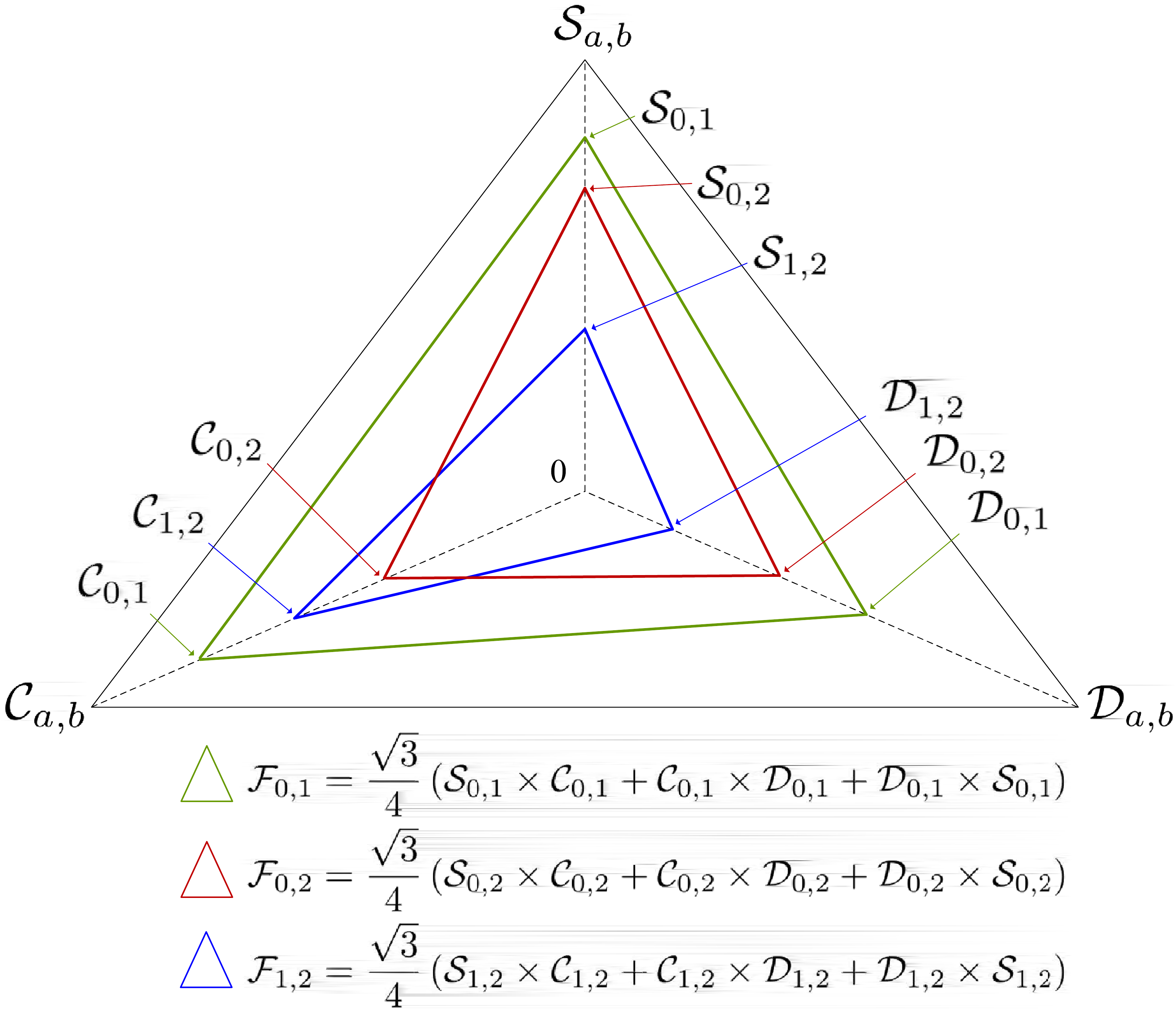}
  \caption{Cluster fusion degree measurement.}
  \label{img0901}
\end{figure}

Based on Definition 4.6, the fusion degree $\mathcal{F}$ between two clusters is expressed as a triangle in an equilateral triangle framework, as shown in Fig. \ref{img0901}.
Vertices of the triangle represent $\mathcal{S}$, $\mathcal{C}$, and $\mathcal{D}$, respectively.
The value of each indicator variable is represented by the segment from the triangle center point to the corresponding vertex.
Then, the value of $\mathcal{F}_{a,b}$ between clusters $c_{a}$ and $c_{b}$ is obtained by calculating the area of the corresponding triangle consisting of $\mathcal{S}_{a,b}$, $\mathcal{C}_{a,b}$, and $\mathcal{D}_{a,b}$, as defined as:
\begin{equation}
\label{eq16}
\begin{aligned}
\mathcal{F}_{a,b} &= S(\mathcal{S}_{a,b}, \mathcal{C}_{a,b}, \mathcal{D}_{a,b}),\\
&= \frac{\sqrt{3}}{4} \left(\mathcal{S}_{a,b} \times \mathcal{C}_{a,b} + \mathcal{C}_{a,b} \times \mathcal{D}_{a,b} +  \mathcal{D}_{a,b} \times \mathcal{S}_{a,b}  \right).
\end{aligned}
\end{equation}
If the value of $\mathcal{F}_{a,b}$ exceeds a given threshold, then clusters $c_{a}$ and $c_{b}$ are merged to a single cluster.
In Fig. \ref{img0901}, there are three triangles with different edge colors, representing the corresponding fusion degrees of three cluster-pairs ($c_{0}, c_{1}, c_{2}$).
Fusion degrees between the merged cluster and other clusters continue to be evaluated.
The process is repeated until the CFDs of all clusters are below the threshold.
An example of cluster fusion degree between three clusters ($c_{0}$ ~ $c_{2}$) is given in Fig. \ref{img0901}.

\begin{figure}[!ht]
  \centering
  \subfigure[Fragmented sub-clusters]{\includegraphics[width=1.6in]{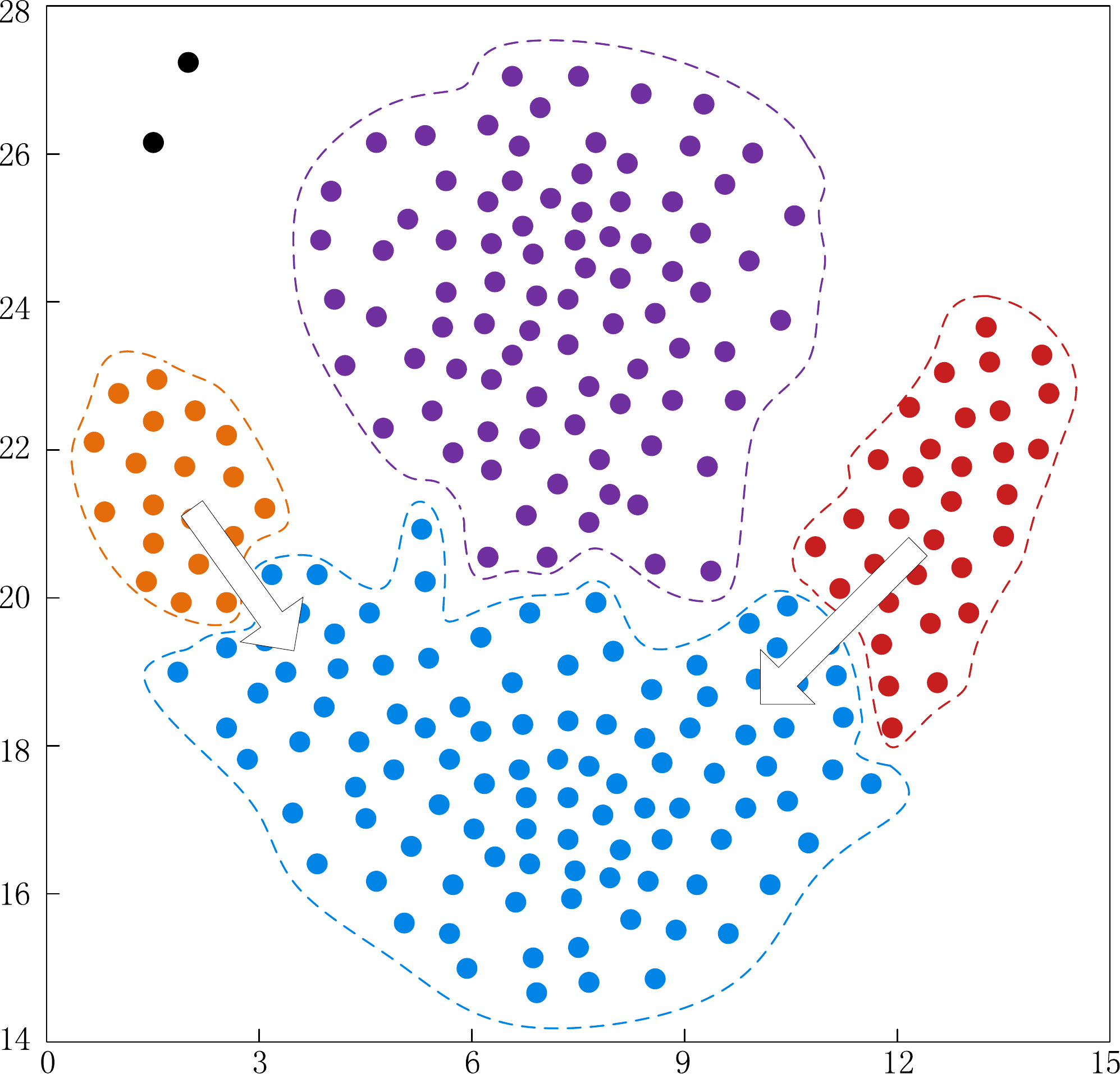}}
  \subfigure[Ensembled clusters]{\includegraphics[width=1.6in]{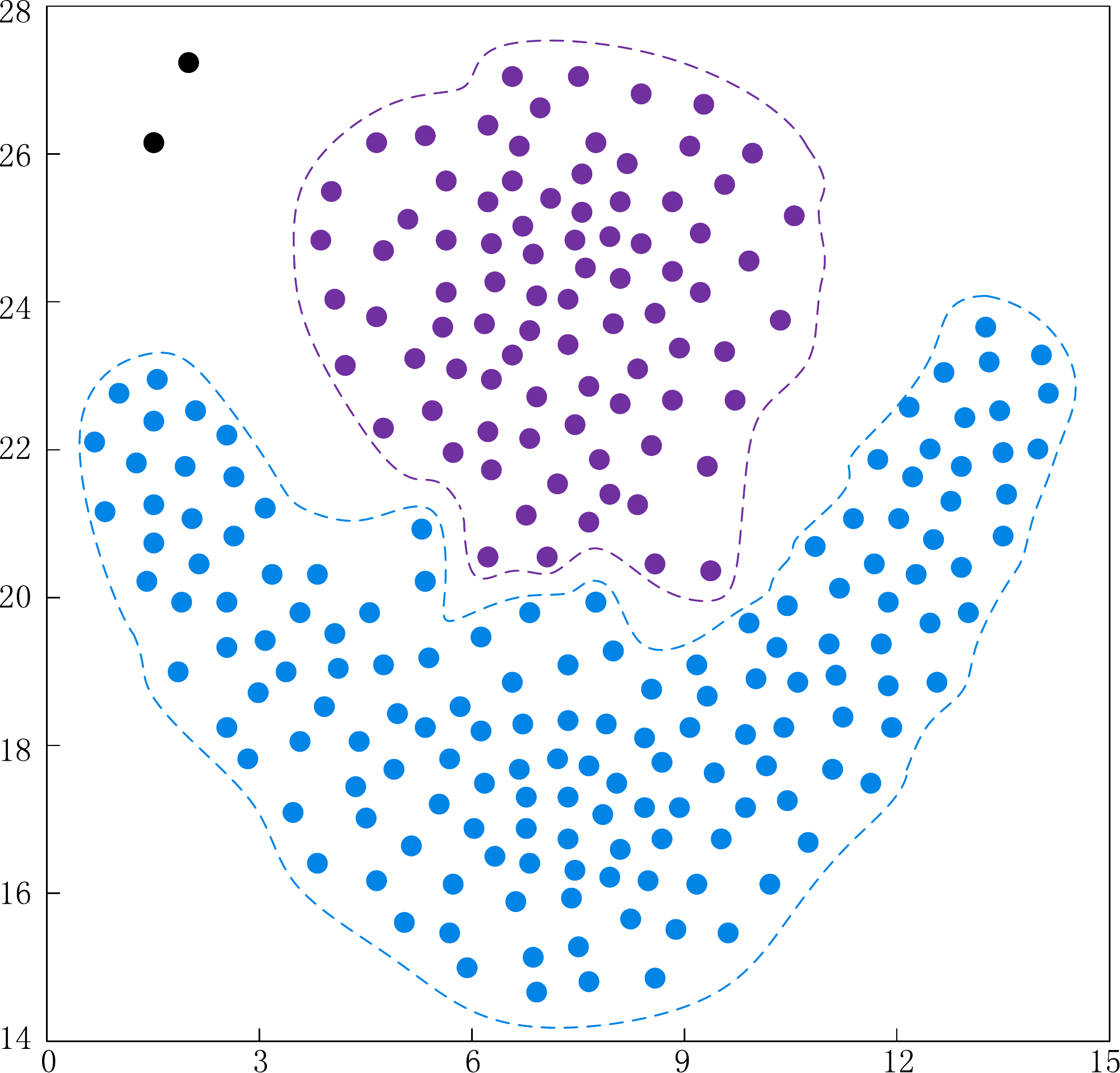}}
  \caption{Example of fragmented cluster ensemble.}
  \label{img10}
\end{figure}

An example of cluster ensemble is shown in Fig. \ref{img10}.
The detailed steps of the cluster self-ensemble process of DADC are presented in Algorithm \ref{alg03}.

\begin{algorithm}[!ht]
\caption{Cluster self-ensemble of DADC.}
\label{alg03}
\begin{algorithmic}[1]
\REQUIRE ~\\
    $IC_{X}$: The initial clusters of $X$;\\
    $\theta_{\mathcal{F}}$: the threshold value of the cluster fusion degree for cluster self-ensemble.\\
\ENSURE ~\\
    $MC_{X}$: the merged clusters of dataset $X$.\\
\WHILE {$IC_{X} \neq \emptyset$}
\STATE get the first cluster $c_{a}$ from $IC_{X}$;
\FOR {each $c_{b} (c_{b} \neq c_{a})$ in $IC_{X}$}
\STATE calculate the inter-cluster density similarity $\mathcal{S}_{a,b}$;
\STATE calculate crossing points $\emph{c}_{(i, a \rightarrow b)}$ and $\emph{c}_{(j, b \rightarrow a)}$;
\STATE calculate cluster crossover degree $\mathcal{C}_{a,b}$;
\STATE calculate cluster density similarity $\emph{d}_{a}$, $\emph{d}_{b}$, and $\emph{d}_{a+b}$;
\STATE calculate cluster density similarity $\mathcal{D}_{a,b}$;
\STATE calculate cluster fusion degree $\mathcal{F}_{a,b}$;
\IF {$\mathcal{F}_{a,b} > \theta_{\mathcal{F}}$}
\STATE  merge clusters $c_{a}' \leftarrow$ merge($c_{a}, c_{b}$);
\STATE  remove $c_{b}$ from $IC_{X}$;
\ENDIF
\ENDFOR
\IF {$c_a = c_{a}'$}
\STATE append $c_{a}$ to the merged clusters $MC_{X}$;
\STATE remove $c_{a}$ from $IC_{X}$;
\ENDIF
\ENDWHILE
\RETURN $MC_{X}$.
\end{algorithmic}
\end{algorithm}

In Algorithm \ref{alg03}, for each initial cluster $c_{a}$ in $IC_{X}$, we respectively calculate the cluster crossover degree, cluster density similarity, and cluster fusion degree between $c_{a}$ and each residual cluster $c_{b}$.
Then, in each iteration, we try to merge the two clusters with the highest cluster fusion degree.
Assuming that the number of initial clusters is $m$, the computational complexity of Algorithm \ref{alg03} is $O(C_{m}^{2})$.

The DADC algorithm consists of processes \ref{alg01}, \ref{alg02}, and \ref{alg03}, requiring the computational complexity of $O(n)$, $O(n)$, and $O(C_{m}^{2})$, respectively.
Thus, the computational complexity of the DADC algorithm is $O(2n + C_{m}^{2})$, where $n$ is the number of points in the dataset $X$ and $m$ is that of the initial clusters.

\section{Experiments}
\label{section5}

\subsection{Experiment Setup}
\label{section5.1}
Experiments are conducted to evaluate the proposed DADC algorithm by comparing with CFSFDP \cite{ex13}, OPTICS \cite{ex85}, DBSCAN \cite{ex15} algorithms in terms of  clustering results analysis and performance evaluation.
The experiments are performed using a workstation equipped with Intel Core i5-6400 quad-core CPU, 8 GB DRAM, and 2 TB main memory.
Two groups of datasets, e.g., synthetic and large-scale real-world datasets, are used in the experiments.
These datasets are downloaded from published online benchmarks, such as the clustering benchmark datasets \cite{ex43} and UCI Machine Learning Repository \cite{ex009}, as shown in Tables \ref{table51} and \ref{table52}.
An implementation of DADC is available from Github at https://github.com/JianguoChen2015/DADC.

\begin{table}[!ht]
\renewcommand{\arraystretch}{1.3}
\setlength{\abovecaptionskip}{0pt}
\setlength{\belowcaptionskip}{0pt}
\caption{Synthetic datasets used in experiments.}
\centering
\small
\label{table51}
\tabcolsep1pt
\begin{tabular}{L{1.5in} C{0.55in} C{0.75in} C{0.5in}}
\hline
 Datasets              &  \#.Samples & \#.Dimensions &  \#.Clusters \\
\hline
  Aggregation          &  350  &  2 &  7 \\
 Compound              &  399  &  2 &  6 \\
 Heartshapes           &  788  &  2 &  3 \\
 Yeast                 &  1484 &  2 &  10\\
 Gaussian clusters (G2)&  2048 &  2 &  2 \\
\hline
\end{tabular}
\end{table}
\begin{table}[!ht]
\renewcommand{\arraystretch}{1.3}
\setlength{\abovecaptionskip}{0pt}
\setlength{\belowcaptionskip}{0pt}
\caption{Large-scale datasets used in experiments.}
\centering
\small
\label{table52}
\tabcolsep1pt
\begin{tabular}{L{1.5in} C{0.55in} C{0.75in} C{0.5in}}
\hline
 Datasets   &  \#.Samples & \#.Dimensions & \#.Clusters \\
\hline
 Individual household electric power consumption (IHEPC) &  275,259 &  9 &  196\\
 Flixster (ASU)                                  &  523,386  &  2 & 153\\
 Heterogeneity activity recognition (HAR) &  930,257 & 16 & 289 \\
 Twitter (ASU)                                   &  316,811 & 2 & 194\\
\hline
\end{tabular}
\end{table}

\subsection{Clustering Results Analysis on Synthetic Datasets}
\label{section5.2}
To clearly and vividly illustrate the clustering results of DADC, multiple experiments are conducted on synthetic datasets in this section by comparing the related clustering algorithms, including DADC, CFSFDP in \cite{ex13}, DBSCAN in \cite{ex15}, and OPTICS in \cite{ex85}.
Synthetic datasets that with the features of VDD, MDDM, and ED, are used in experiments.

\subsubsection{Clustering Results on VDD Datasets}
To illustrate the effectiveness of the proposed method of the domain-adaptive density measurement in DADC, we conduct experiments on VDD datasets.
Fig. \ref{img01} (a) is a synthetic dataset (\emph{Heartshpes}) described in Table \ref{table51}, which is composed of three heart-shaped regions with different densities.
Each region contains 71 data points.

\begin{figure}[!ht]
  \centering
  \subfigure[Data points]{\includegraphics[width=1.6in]{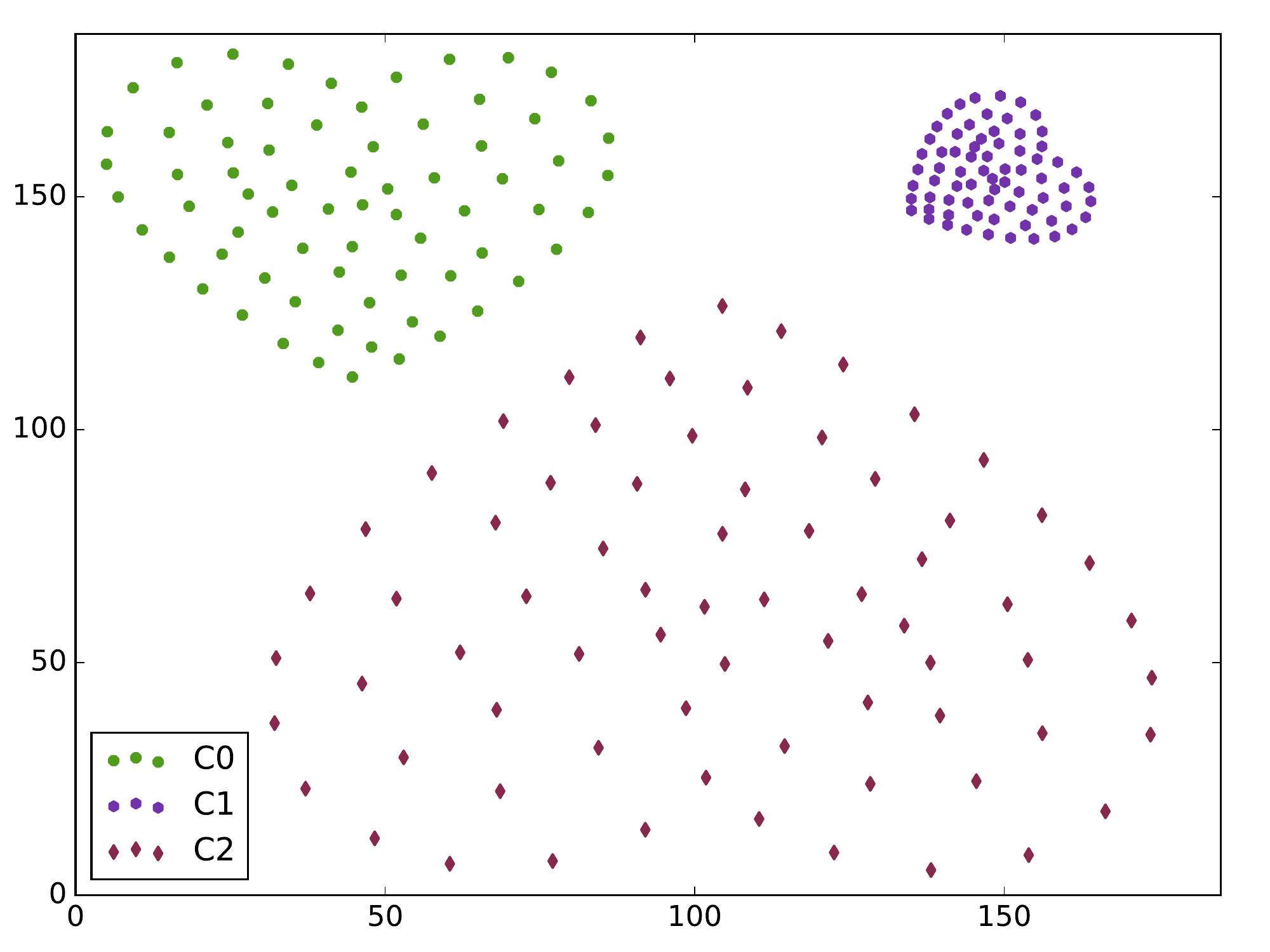}}
  \subfigure[Local/domain density]{\includegraphics[width=1.6in]{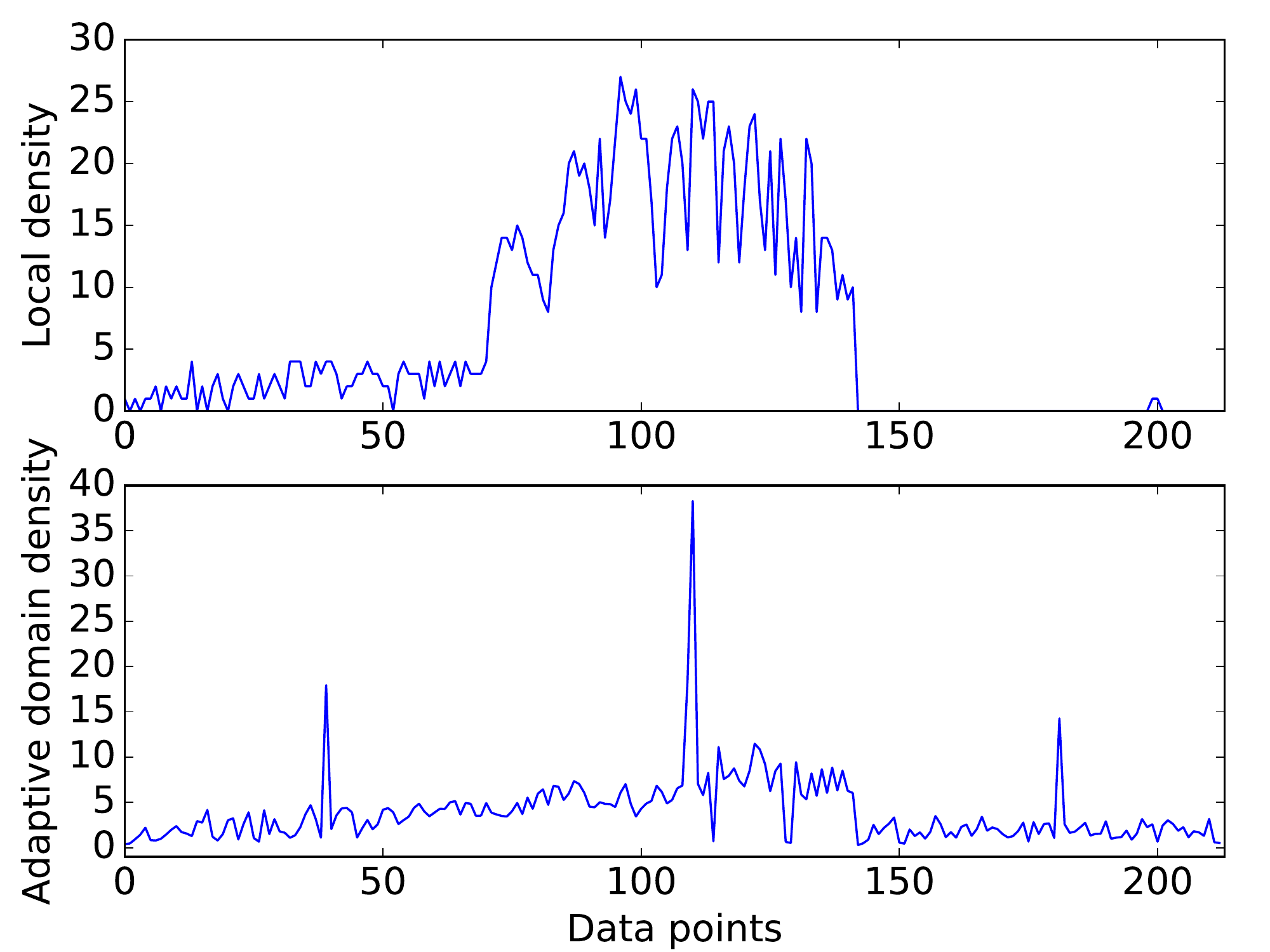}}
  \subfigure[Decision graph of CFSFDP]{\includegraphics[width=1.6in]{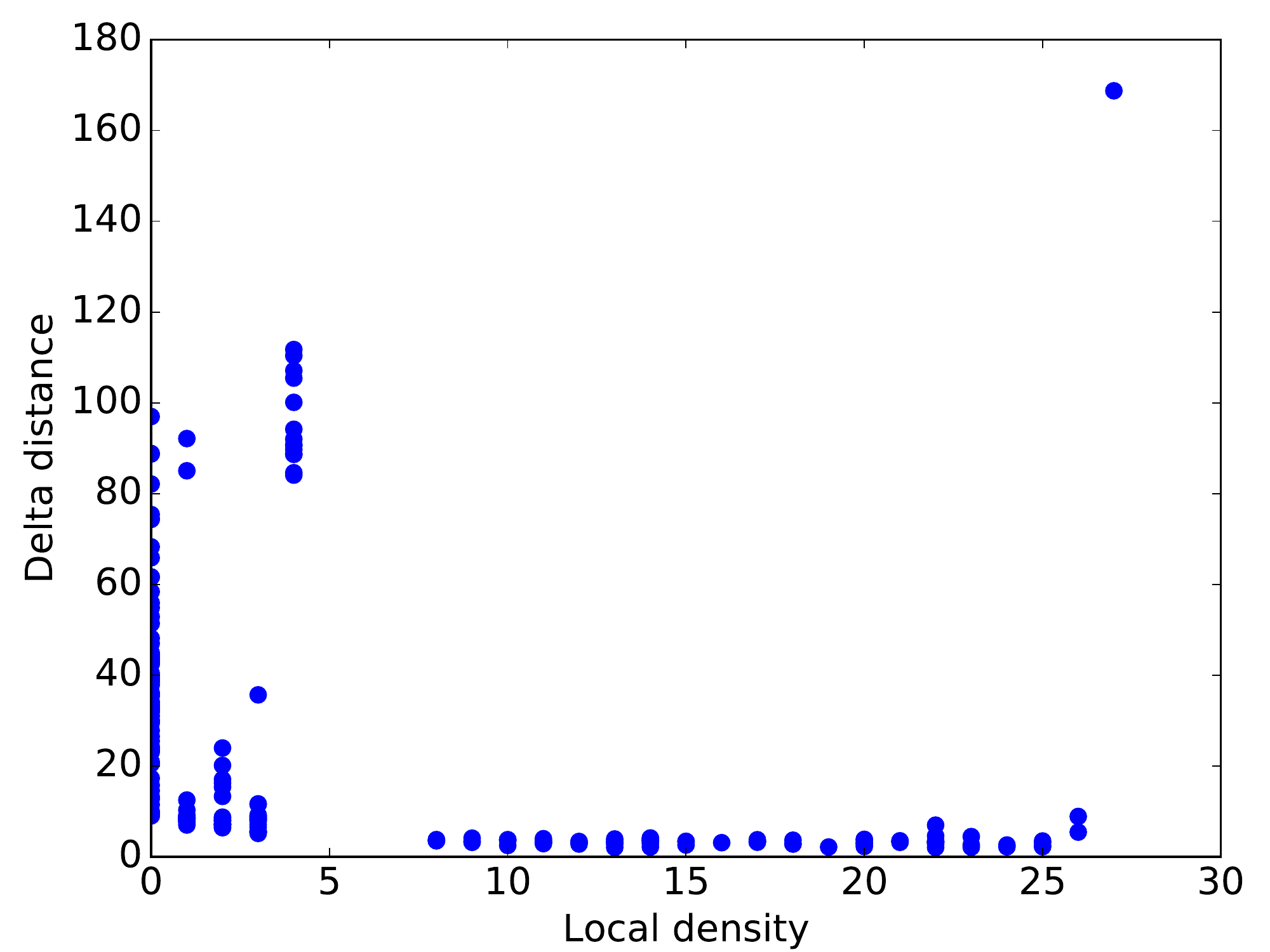}}
  \subfigure[Decision graph of DADC]{\includegraphics[width=1.6in]{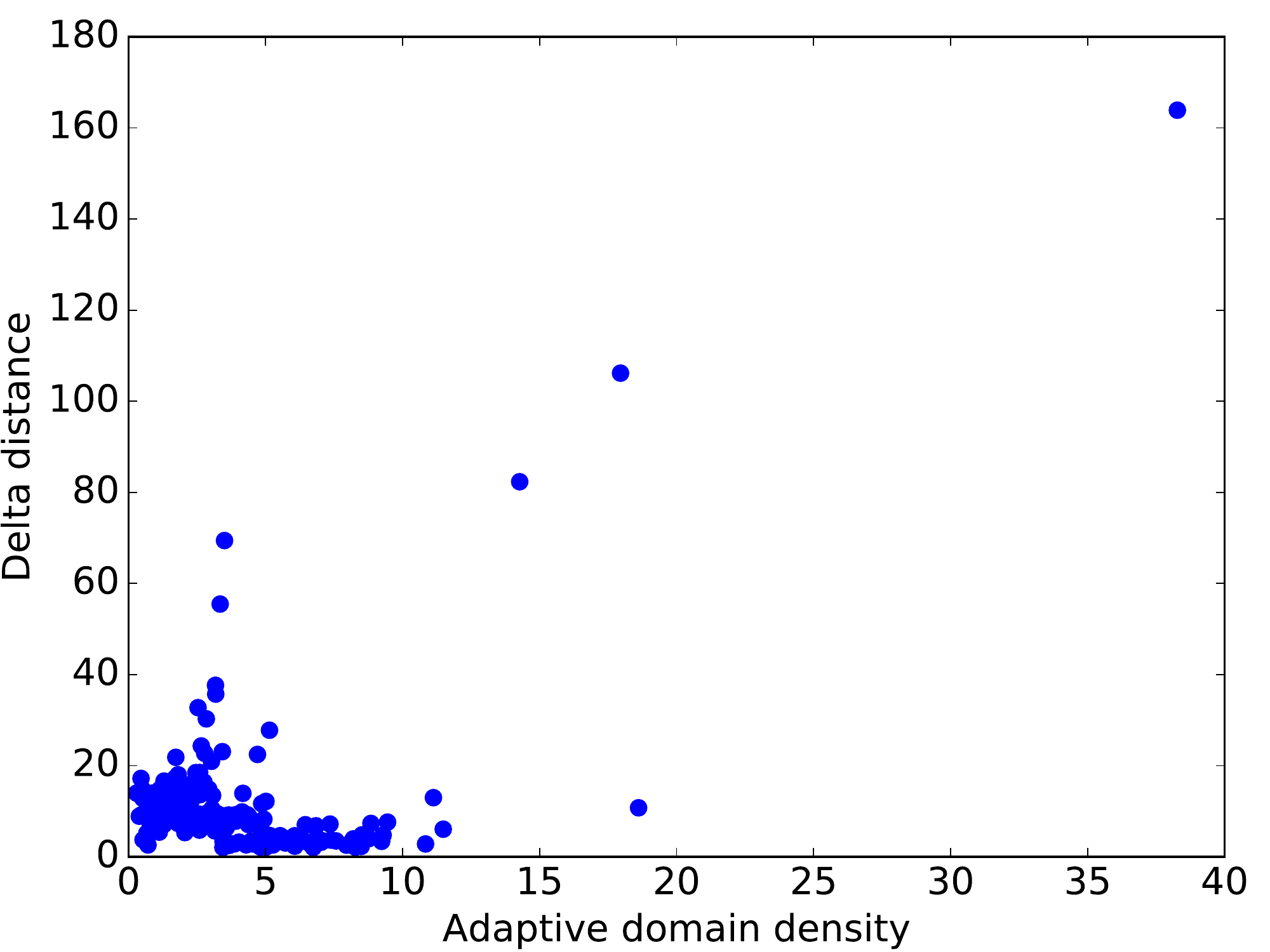}}
  \caption{Decision graphs for VDD dataset.}
  \label{img11}
\end{figure}

The local density and domain density of each data point are calculated by the original measurement of CFSFDP and the domain adaptive measurement of DADC separately, as shown in Fig. \ref{img11} (b).
It is evident from Fig. \ref{img11} (b), according to CFSFDP, the local densities of data points in the second region (no. 72-142) are far higher than that of the other two relatively sparse regions.
In this case, it has difficulty in detecting the local-density peaks in the sparse regions.
In contrast, according to DADC, although the density distribution of the three regions are different, the domain-density peaks of each region are obviously identified.
As shown in Fig. \ref{img01} (b), there is only one decision point in the CFSFDP clustering decision graph that is detected as a cluster center.
More than 140 points have low values of local density and high values of Delta distance and are detected as outliers.
In contrast, in the DADC clustering decision graph of Fig. \ref{img11} (d), there are three decision points with high values of both domain-adaptive density and Delta distance, which are identified as the cluster centers of three regions separately.
The clustering results show that the proposed DADC algorithm can effectively detect the domain-density peaks of data points and identify clusters in different density regions.

\begin{figure}[!ht]
  \centering
  \subfigure[Data points of Aggregation]{\includegraphics[width=1.6in, height=1.6in]{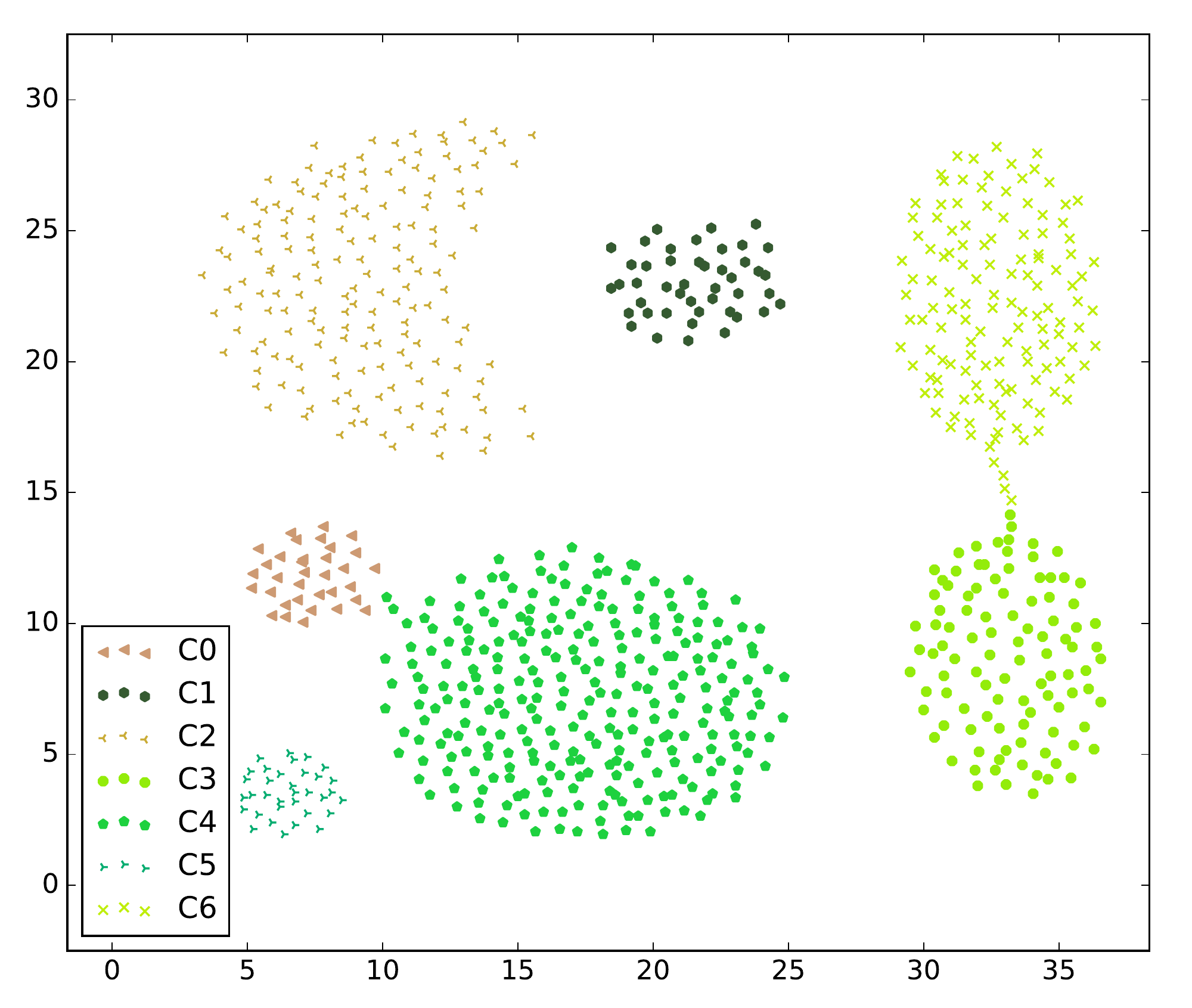}}
  \subfigure[Local/domain densities on Aggregation]{\includegraphics[width=1.6in]{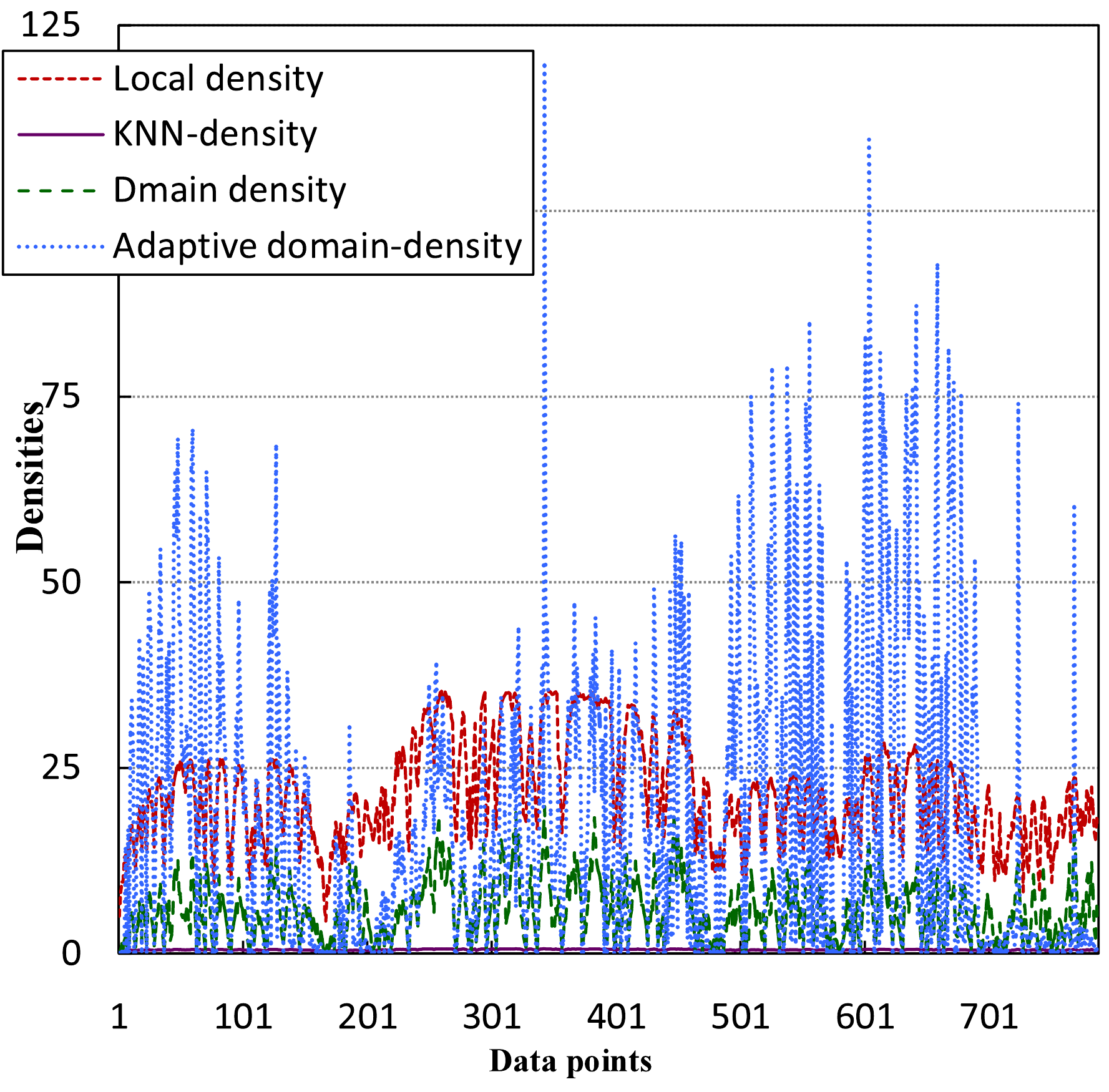}}
  \subfigure[Data points of Compound]{\includegraphics[width=1.6in, height=1.6in]{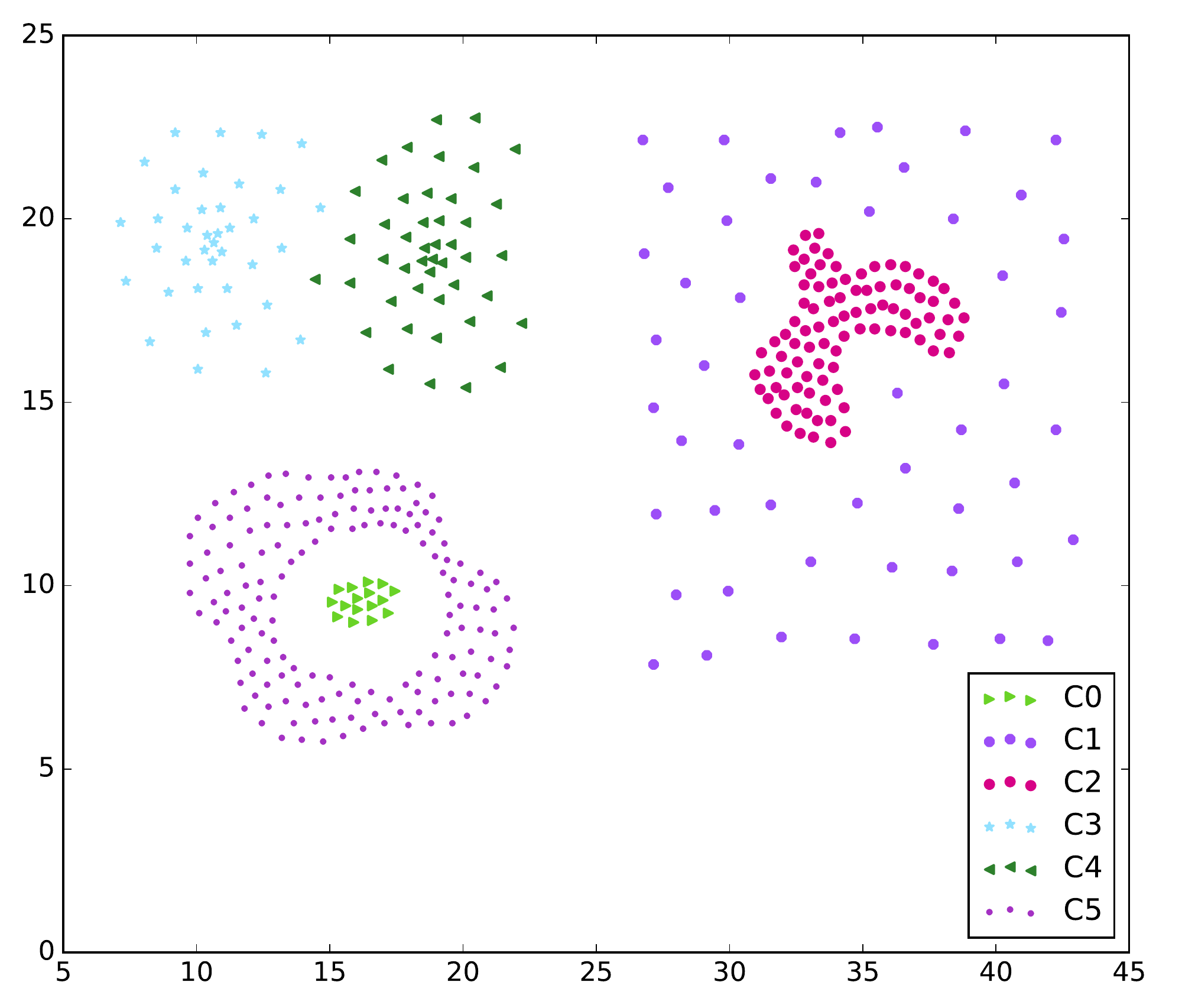}}
  \subfigure[Local/domain densities on Compound]{\includegraphics[width=1.6in]{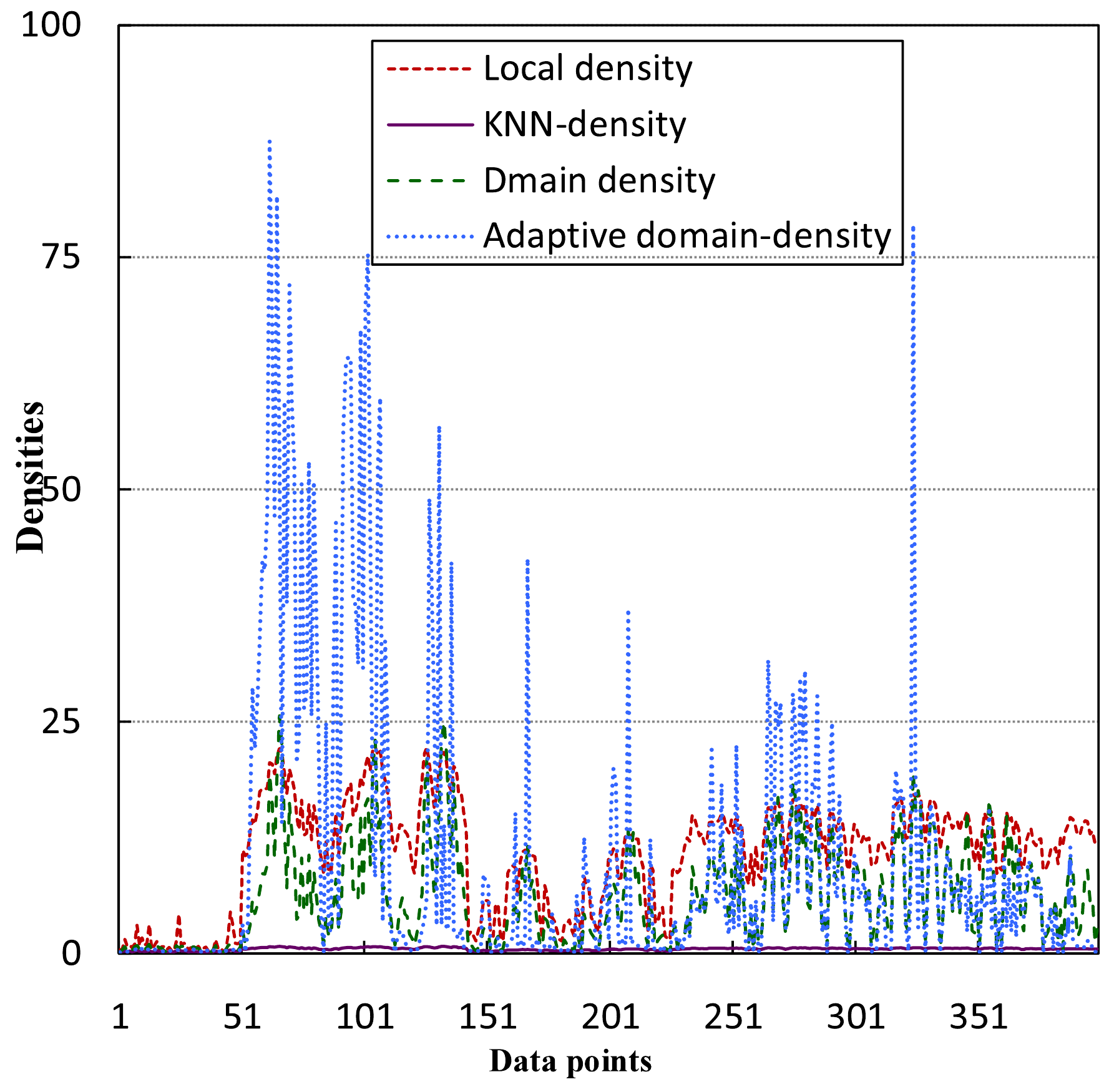}}
  \caption{Adaptive domain-densities on VDD datasets.}
  \label{img12}
\end{figure}

Two groups of VDD datasets (\emph{Aggregation} and \emph{Compound}) are used in the experiments to further evaluate the clustering effectiveness of DADC by comparing the CFSFDP algorithm.
The local density of each data point is obtained by the CFSFDP algorithm, while the KNN-density, domain density, and domain-adaptive density of each data point are calculated by the proposed DADC algorithm.
The comparison results are illustrated in Fig. \ref{img12}.

As shown in Fig. \ref{img12} (b) and Fig. \ref{img12} (d), the local densities of data points in dense regions are obviously higher than those of data points in sparse regions.
It is easy to treat the data points in sparse regions as noisy data rather than independent clusters.
In contrast, by the method of DADC, domain-adaptive densities of all of the data points are detected with obvious differences.
Although the datasets have multiple regions with different densities, the domain-density peaks in each region are quickly identified.
More comparison results on VDD datasets are described in supplementary material.

\subsubsection{Clustering Results on ED Datasets}
To evaluate the effect of the cluster self-identification method of the proposed DADC algorithm, experiments are performed on an ED dataset (\emph{Hexagon)} by comparing to CFSFDP, OPTICS, and DBSCAN algorithms, respectively.
The clustering results are shown in Fig. \ref{img13}.
More comparison results on ED datasets are described in supplementary material.

\begin{figure}[!ht]
  \centering
  \subfigure[DADC on Hexagon]{\includegraphics[width=1.6in]{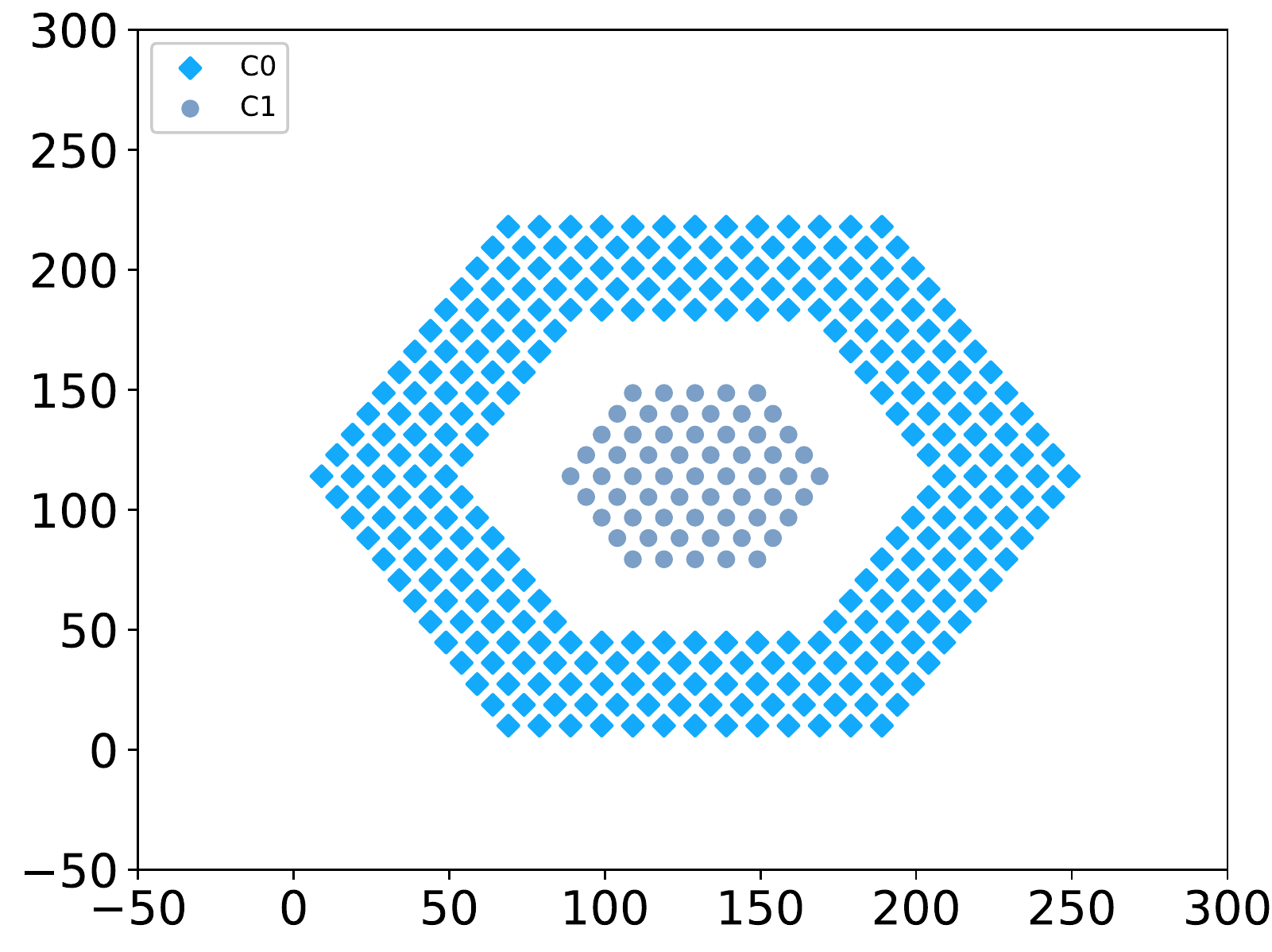}}
  \subfigure[CFSFDP on Hexagon]{\includegraphics[width=1.6in]{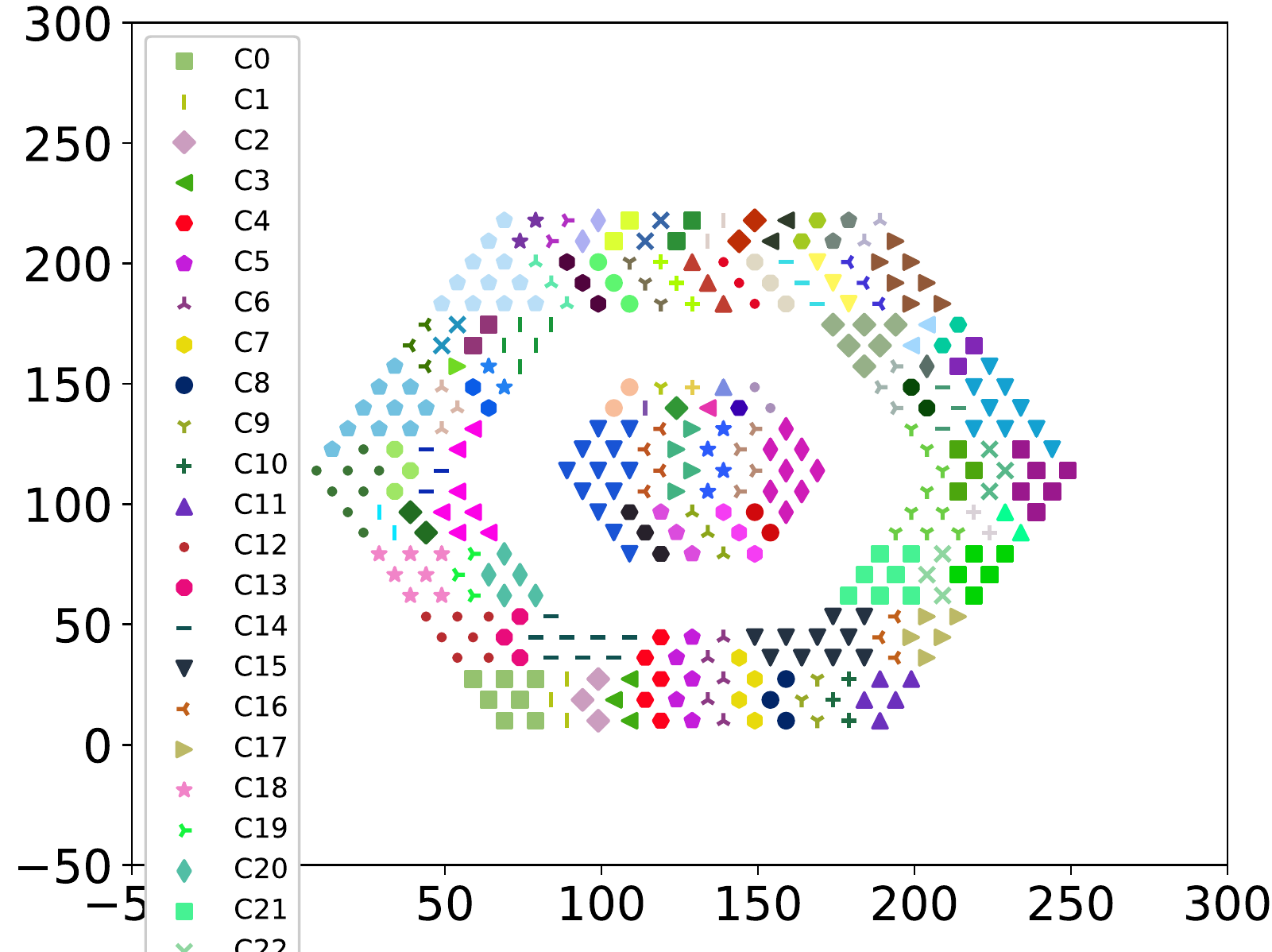}}
  \subfigure[OPTICS on Hexagon]{\includegraphics[width=1.6in]{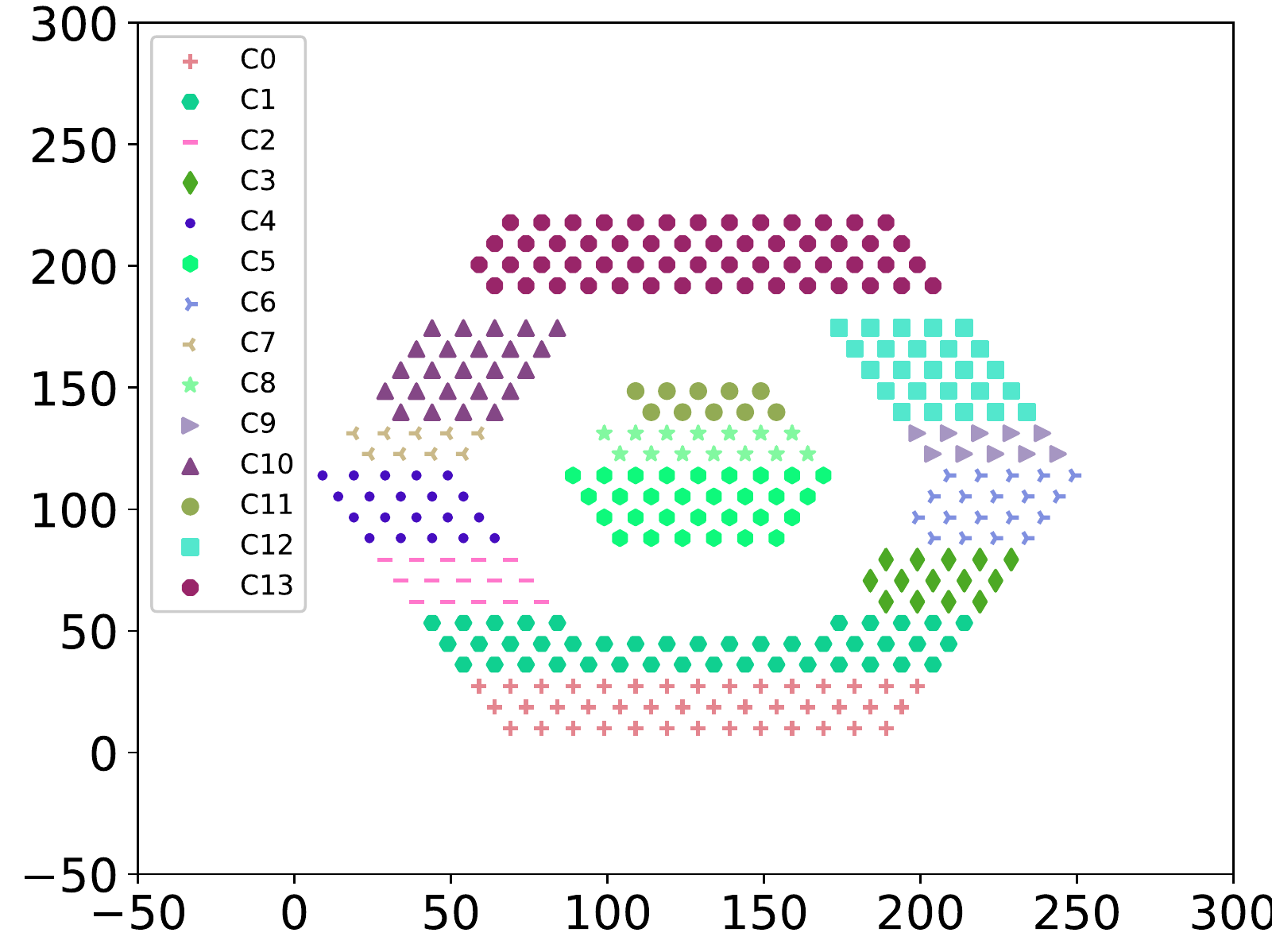}}
  \subfigure[DBSCAN on Hexagon]{\includegraphics[width=1.6in]{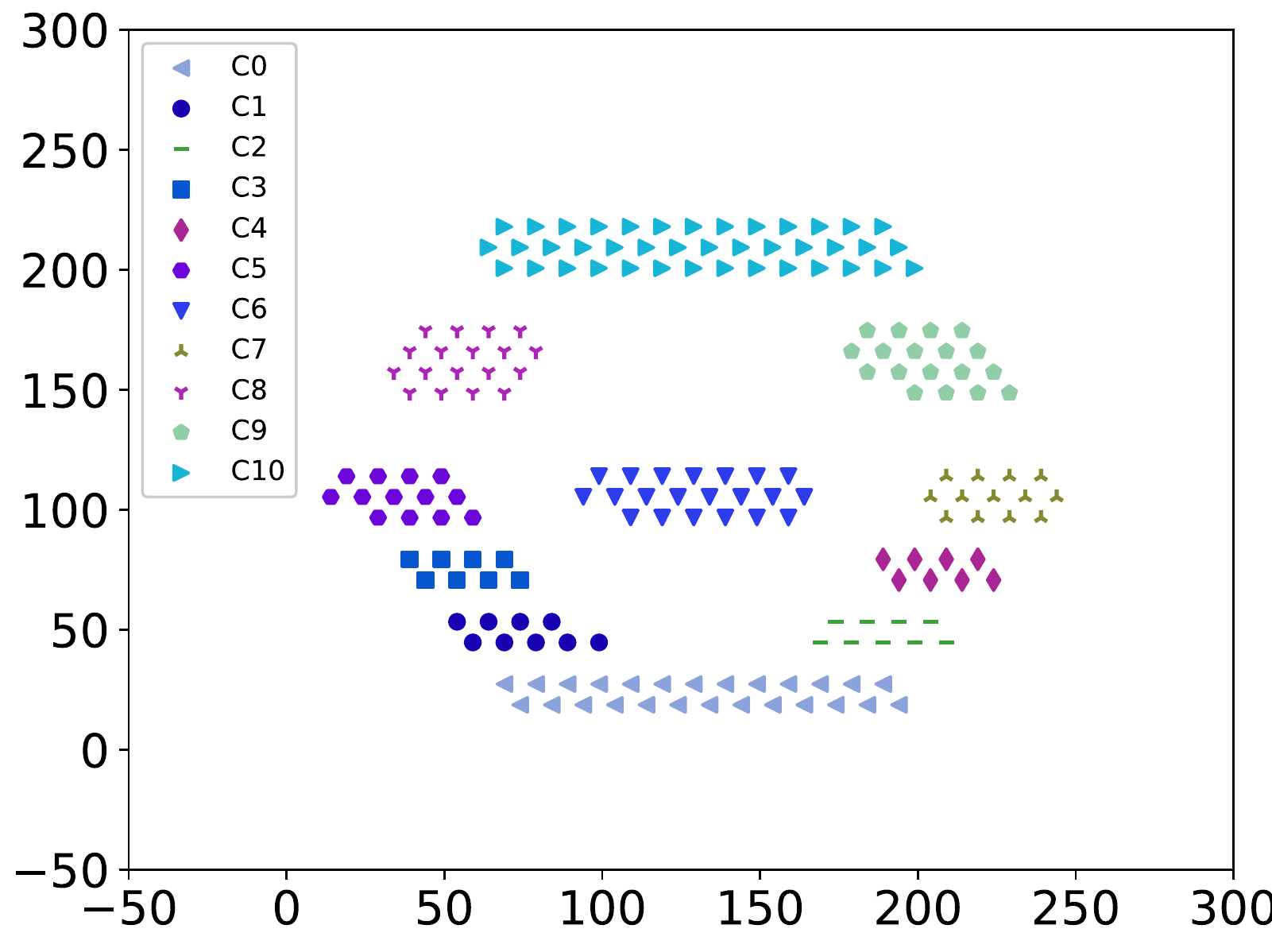}}
  \caption{Clustering results on equilibrium distributed datasets.}
  \label{img13}
\end{figure}

Since the ED dataset does not have local-density peaks, CFSFDP obtains numerous fragmented clusters.
Making use of the cluster self-identification and cluster ensemble process, the proposed DADC algorithm can merge the fragmented clusters effectively.
Therefore, DADC effectively solves this problem and obtains accurate clustering results.
As shown in Fig. \ref{img13} (a), the dataset is clustered into two clusters by DADC.
In contrast, as shown in Fig. \ref{img13} (b) - (d), more than 231 fragmented clusters are produced by CFSFDP.
Clustering results of OPTICS and DBSCAN are very sensitive to the parameter thresholds of eps (connectivity radius) and minpts (minimum number of shared neighbors).
For example, for both OPTICS and DBSCAN, we set their parameter thresholds of eps and minpts to 10 and 5, and generate 14 and 11 fragmented clusters by OPTICS and DBSCAN, respectively.
The experimental results show that the proposed DADC algorithm is more accurate than other algorithms for ED datasets clustering.

\subsubsection{Clustering Results on MDDM Datasets}
For datasets with MDDM characteristics, multi domain-density maximums might lead to fragmented clusters, whose overall distribution is similar to that of adjacent clusters.
We conduct comparison experiments on a dataset with MDDM characteristics to evaluate the clustering effect of the comparative clustering algorithms.
A synthesized dataset (G2) with MDDM characteristics is used in the experiments. The clustering results are shown in \ref{img14}.
More comparison results on MDDM datasets are described in supplementary material.

\begin{figure}[!ht]
  \centering
  \subfigure[DADC on G2]{\includegraphics[width=1.6in]{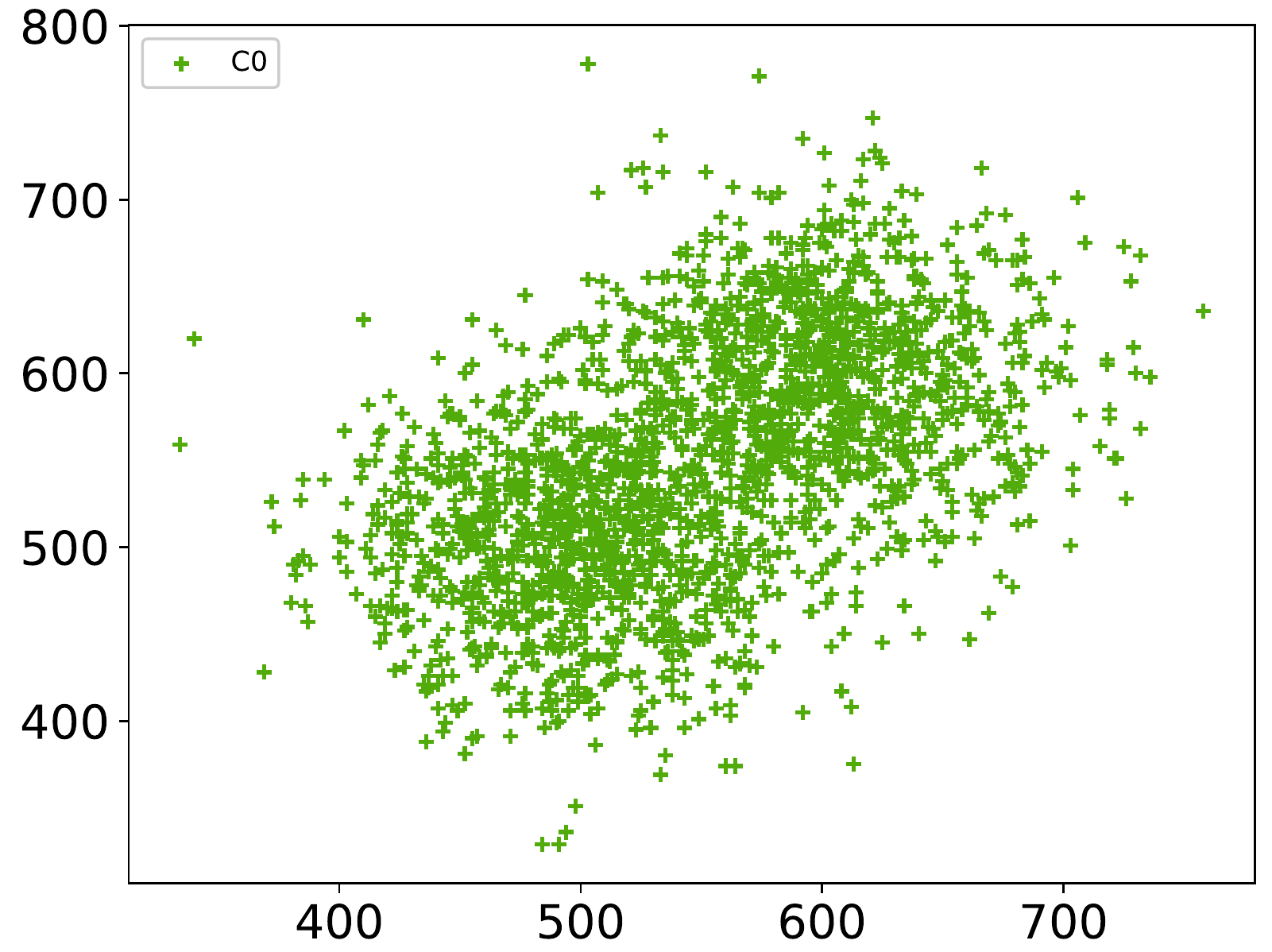}}
  \subfigure[CFSFDP on G2]{\includegraphics[width=1.6in]{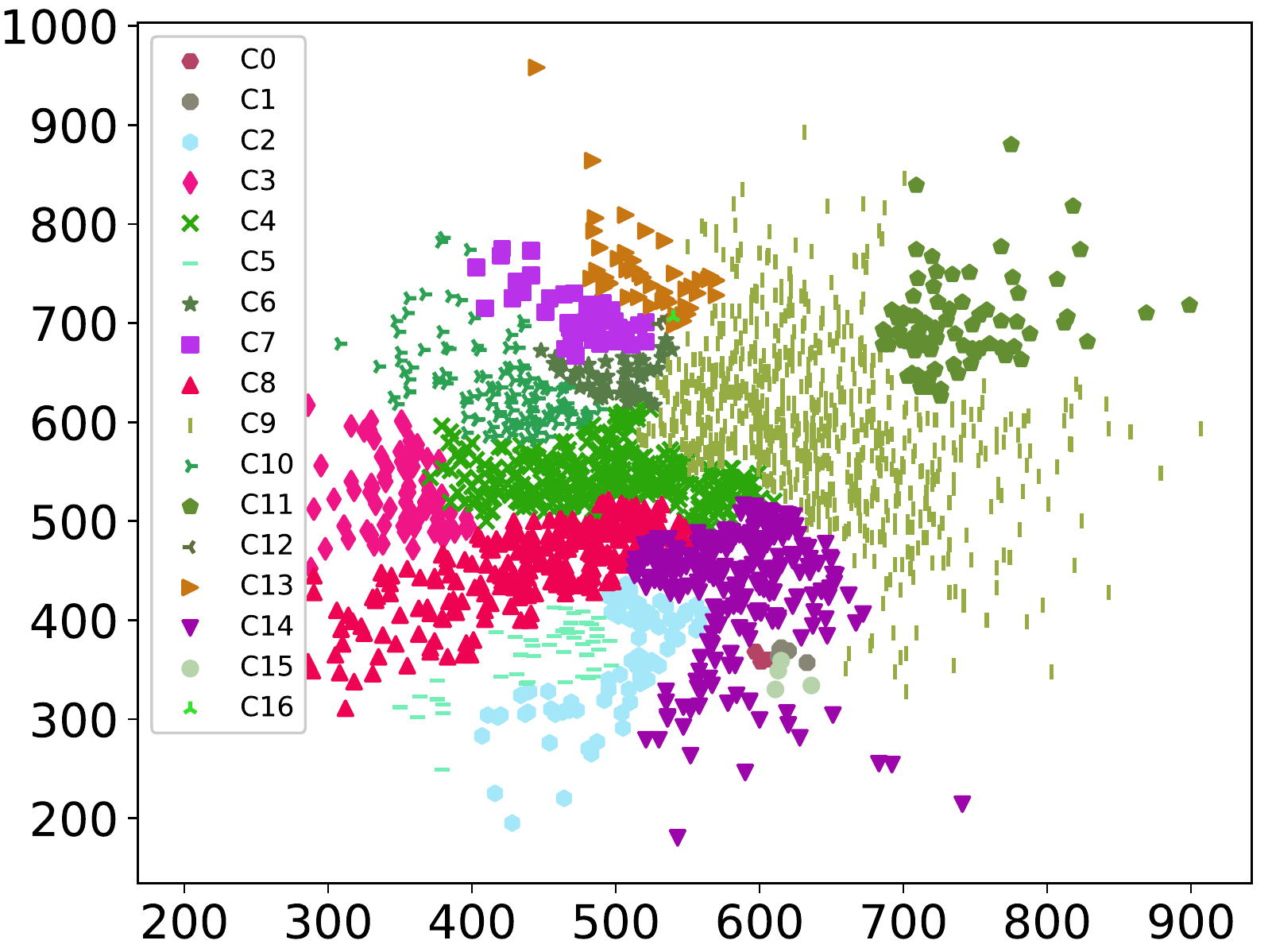}}
  \subfigure[OPTICS on G2]{\includegraphics[width=1.6in]{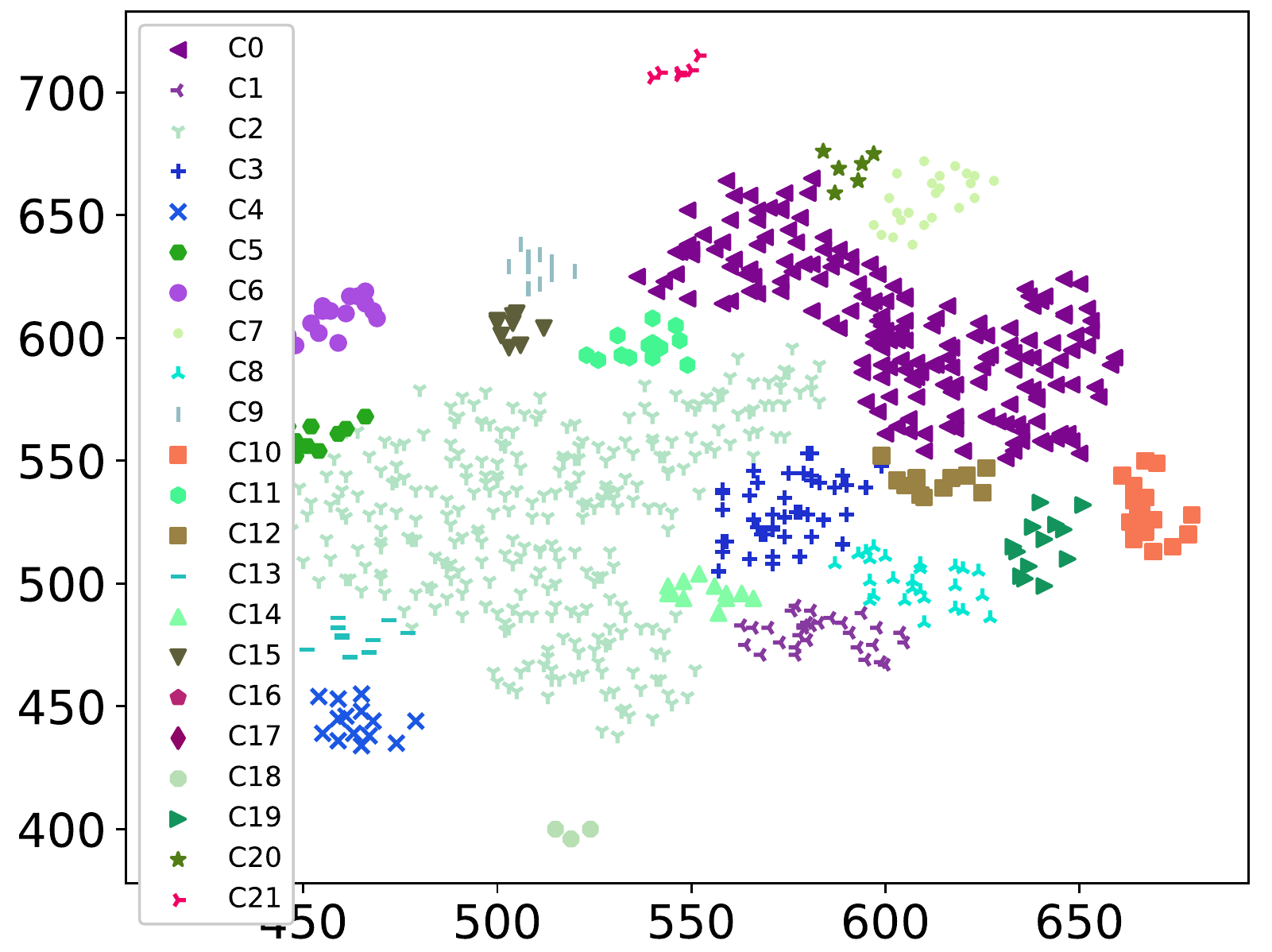}}
  \subfigure[DBSCAN on G2]{\includegraphics[width=1.6in]{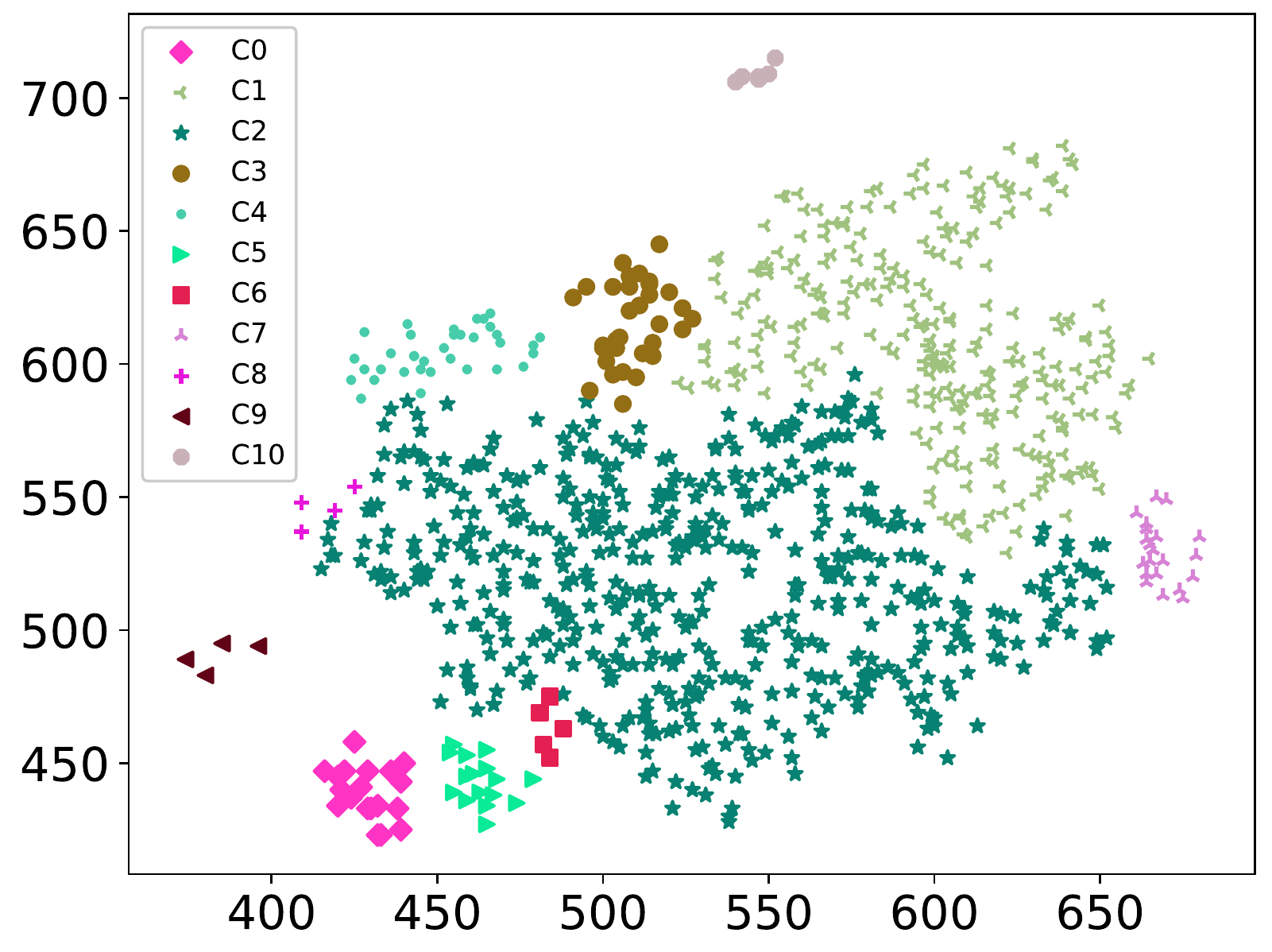}}
  \caption{Clustering results on MDDM dataset.}
  \label{img14}
\end{figure}

After obtaining 17 density peaks using CFSFDP, 17 corresponding clusters are generated as shown in Fig. \ref{img14} (b).
However, these clusters have similar overall density distribution, and it is reasonable to merge them into a single cluster.
DADC can eventually merge the 17 fragmented clusters into one cluster, as shown in Fig. \ref{img14} (a).
Again, the clustering results of OPTICS and DBSCAN are very sensitive to the parameter thresholds of eps and minpts.
As shown in Fig. \ref{img14} (c) and (d), when we set the OPTICS and DBSCAN algorithms' parameter thresholds of eps and minpts to 13 and 10, then, 22 and 11 fragmented clusters are clustered by OPTICS and DBSCAN, respectively.
Compared with CFSFDP, OPTICS, and DBSCAN, experimental results show that DADC achieves more reasonable clustering results on MDDM datasets.

\subsection{Performance Evaluation}
\label{section5.3}

\subsubsection{Clustering Accuracy Analysis}
Clustering Accuracy (CA) is introduced to evaluate the clustering algorithms.
CA measures the ratio of the correctly classified/clustered instances to the pre-defined class labels.
Let $X$ be the dataset in this experiment, $C$ be the set of classes/clusters detected by the corresponding algorithm, and $L$ be the set of pre-defined class labels.
CA is defined in Eq. (\ref{eq17}):

\begin{equation}
\label{eq17}
CA = \sum_{i=0}^{K-1}{\frac{\max{(C_{i}|L_{i})}}{|X|}},
\end{equation}
where $C_{i}$ is the data points in the $i$-th class/cluster, $L_{i}$ is the pre-defined class labels of the data points in $C_{i}$, and $K$ is the number of $C$.
$max(C_{i}|L_{i})$ is the number of data points that have the majority label in $C_{i}$.
The greater value of CA means that the higher accuracy of the classification / clustering algorithm, and each cluster achieves high purity.
The experimental results of the clustering accuracy comparison are given in Table \ref{table53}.

\begin{table}[!ht]
\renewcommand{\arraystretch}{1.3}
\caption{Clustering accuracy comparison.}
\centering
\small
\label{table53}
\tabcolsep1pt
\begin{tabular}{L{0.9in} C{0.6in} C{0.55in} C{0.5in} C{0.5in}}
\hline
 Datasets       &  DADC         &  CFSFDP       &   OPTICS      &  DBSCAN \\
\hline
 Heartshapes	&  100.00\%     &	 83.42\%	&    91.33\%	&    91.33\% \\
 Yeast	        &    91.67\%	&    83.23\%    &	 82.54\%	&  80.41\% \\
 G2	            &    100.00\%	&     90.45\%   &	 84.23\%	&  82.85\% \\
\hline
 IHEPC	        &    92.34\%	&  87.72\%	    &    73.98\%	&    62.03\% \\
 Flixster	    &    87.67\%	&   79.09\%	    &    65.31\%	&    55.51\% \\
 Twitter	    &    72.26\%	&    68.85\%	&    53.90\%	&    51.42\% \\
 HAR	        &    83.29\%	&    84.23\%	&    58.26\%	&    56.92\% \\
\hline
\end{tabular}
\end{table}

As shown in Table \ref{table53}, DADC outperforms others on both synthetic and large-scale real-world datasets.
In the case of Friendster, the average CA of DADC is 87.67\%, while that of CFSFDP is 79.09\%, that of OPTICS is 65.31\%, and that of DBSCAN is 55.51\%.
For synthetic datasets, DADC achieves a high average CA of 97.22\%.
The average accuracies of OPTICS and DBSCAN algorithms are noticeably lower than that of CFSFDP and DADC.
For the large-scale real-world datasets, CA of DADC is higher than that of the compared algorithms, keeping in the range of 72.26\% and 92.34\%.
It illustrates that DADC achieves higher clustering accuracy over CFSFDP, OPTICS, and DBSCAN algorithms.

\subsubsection{Robustness Analysis}
\begin{figure}[!ht]
  \centering
  \includegraphics[width=2.5in]{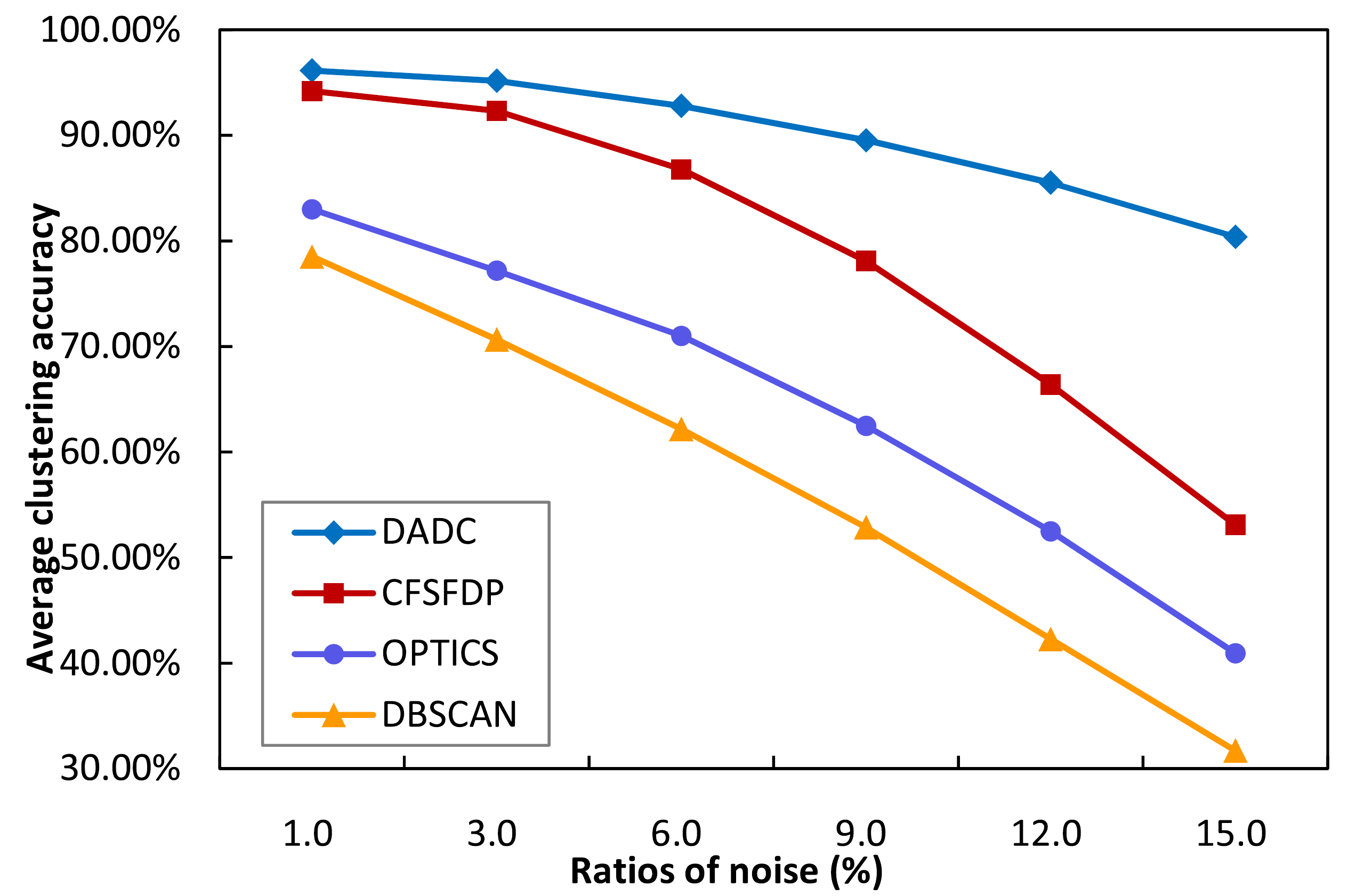}
  \caption{Comparison of algorithm robustness.}
  \label{img15}
\end{figure}

Experiments are conducted to evaluate the robustness of the compared algorithms on noisy datasets.
Four groups of real-world datasets from practical applications described in Table \ref{table52} are used in the experiments with different degrees of noise.
We generate different amounts of random and non-repetitive data points as noise in the value space of the original dataset.
The noise-level of each dataset gradually increases from 1.0\% to 15.0\%.
The experimental results are presented in Fig. \ref{img15}.

As observed in Fig. \ref{img15}, with the increasing proportion of noise, the average clustering accuracy of each algorithm decreases, respectively.
However, the average clustering accuracy of DADC drops at a minimum rate, while those of CFSFDP and OPTICS lie in the second and third, while that of DBSCAN declines at the fastest speed.
When the noise-level rises from 1.0\% to 15.0\%, the average accuracy of DADC decreases from 96.14\% to 80.39\%, which indicates that DADC is most robust to different noise-level data.
The average accuracy of CFSFDP drops from 94.21\% to 53.18\%, and that of DBSCAN decreases from 78.52\% to 31.74\%.
For example, when the scale of noisy data increases from 1.0\% to 15.0\%, the average clustering accuracy of DADC reduces from 96.0\% to 80.3\%.
Compared to the compared algorithms, DADC retains higher accuracy in each case.
Therefore, DADC illustrates higher robustness than compared algorithms to noisy data.

\section{Conclusions}
\label{section6}
This paper presented a domain-adaptive density clustering algorithm, which is effective in the datasets with varying-
density distribution (VDD), multiple domain-density maximums (MDDM), or equilibrium distribution (ED).
A domain adaptive method was proposed to calculate domain densities and detect density peaks of data points in VDD datasets, and cluster centers were identified.
In addition, a cluster fusion degree model and a CFD-based cluster self-ensemble method were proposed to merge fragmented clusters with minimum artificial intervention in MDDM and ED datasets.
In comparison with existing clustering algorithms, the proposed DADC algorithm requires fewer parameters and non-iterative nature, achieving outstanding advantages in terms of accuracy and robustness.

As future work, we will further research issues of big data clustering analysis, including incremental clustering, time-series data clustering, and parallel clustering in distributed and parallel computing environments.

\bibliographystyle{IEEEtran}
\bibliography{reference}

\end{document}